\documentclass[a4paper,fleqn]{cas-dc}

\usepackage[authoryear]{natbib}
\usepackage{amsmath,amssymb,mathtools}
\usepackage{booktabs}
\usepackage{graphicx}
\usepackage{placeins}
\usepackage{float}
\usepackage{capt-of}
\usepackage{microtype}
\usepackage{xcolor}
\usepackage{enumitem}
\usepackage{url}


\newcommand{\method}{HiGDiff}
\newcommand{\R}{\mathbb{R}}
\newcommand{\sg}{\operatorname{sg}}
\newcommand{\render}{\mathcal{R}_{\mathrm{G}}}
\newcommand{\project}{\mathcal{P}_{\Theta}}
\newcommand{\backproject}{\mathcal{B}_{\Theta}}
\newcommand{\analytic}{\mathcal{A}_{\Theta}}
\newcommand{\N}{\mathcal{N}}

\ExplSyntaxOn

\ExplSyntaxOff

\begin{document}

\let\WriteBookmarks\relax
\def\floatpagepagefraction{1}
\def\textpagefraction{.001}

\shorttitle{HiGDiff}
\shortauthors{Yang et al.}

\title[mode=title]{Feed-Forward Hierarchical Gaussian Diffusion for Extreme CT Reconstruction}

\author[inst1,inst2]{Yuezhe Yang}
\ead{yuezhe1@ualberta.ca}
\author[inst1]{Li Cheng}
\cormark[1]
\ead{lcheng5@ualberta.ca}
\cortext[cor1]{Corresponding author}

\affiliation[inst1]{organization={Department of Electrical and Computer Engineering},
    addressline={University of Alberta},
    city={Edmonton},
    state={Alberta},
    postcode={T6G 1H9},
    country={Canada}}
\affiliation[inst2]{organization={School of Computer Science},
    addressline={University of Sydney},
    city={Sydney},
    state={New South Wales},
    postcode={2006},
    country={Australia}}

\begin{abstract}
Reconstructing three-dimensional computed tomography (CT) from severely constrained projections is highly ill-posed. Sparse angular sampling, restricted angular coverage, and low photon counts can occur individually or jointly, obscuring global anatomy and local tissue detail. Many learned CT reconstruction methods are tailored to a single dominant degradation. Existing diffusion and Gaussian approaches commonly recover global structure and local detail within a shared representation. We propose \method{}, a feed-forward hierarchical Gaussian diffusion framework that decomposes reconstruction both spatially and from structure to detail. Physics-conditioned anatomical anchors and a foreground capacity field allocate learnable Gaussian primitives to informative regions. A structure diffusion stage first recovers global attenuation geometry, and its learned representation conditions a detail diffusion stage for residual boundaries and tissue transitions. The resulting Gaussian banks are rendered as attenuation fields and further refined by a gradient-isolated residual module. Experiments on three distinct CT benchmark datasets demonstrate state-of-the-art reconstruction performance across isolated, paired, and joint degradation settings, including improvements of 5.81 dB in macro-average peak signal-to-noise ratio (PSNR) and 0.113 in structural similarity index measure (SSIM) on the Low Dose CT Image and Projection Data (LDCT-PD) collection. Code and experimental configurations are openly available at \href{https://github.com/Bean-Young/HiGDiff}{github.com/Bean-Young/HiGDiff}.
\end{abstract}


\maketitle

\section{Introduction}
\label{sec:introduction}

Computed tomography (CT) reconstructs a volumetric attenuation field from X-ray line integrals and supports quantitative tissue assessment, lesion localization, and three-dimensional (3D) anatomical visualization \citep{mayo1991high,adams2009quantitative}. Three acquisition constraints degrade the measurements through different mechanisms. Sparse-view acquisition reduces angular sampling density and produces globally distributed streaks and angular-aliasing artifacts. Limited-angle acquisition restricts angular coverage, removes direction-dependent information, and causes anisotropic blurring and characteristic slope artifacts \citep{barrett2004artifacts}. Low-dose acquisition lowers the photon flux and increases signal-dependent Poisson noise in the measured projections \citep{liang2017guest}. We study severely constrained CT acquisition across individual, paired, and joint degradation settings. These conditions reduce the available projection evidence through one or more acquisition factors. Joint degradation couples sampling aliasing, directional incompleteness, and photon noise, obscuring both long-range anatomy and local tissue transitions. Successful reconstruction must recover anatomical continuity, preserve subtle detail, and prevent acquisition artifacts from appearing as anatomical structures across this spectrum.

Analytical and iterative reconstruction expose a persistent trade-off between accuracy and efficiency. Filtered back projection (FBP) offers high speed and degrades sharply with incomplete or noisy measurements. Model-based iterative reconstruction incorporates the acquisition operator and handcrafted regularization at a considerably higher computational cost \citep{deak2013filtered,koetzier2023deep}. Learned post-processing and unrolled networks integrate data-driven priors with analytic or iterative inversion, as demonstrated by RED-CNN, FBPConvNet, and LEARN \citep{chen2017redcnn,jin2017deep,chen2018learn}. Frequency-aware and dual-domain methods further target sparse-view artifacts and projection noise \citep{ma2023freeseed,li2025ddoct}. These approaches are commonly configured around one dominant degradation or a selected pair. Their dense voxel parameterizations assign representation sites uniformly across the volume, which constrains anatomy-aware allocation of reconstruction capacity under severely limited measurements.

Diffusion priors provide expressive image distributions and integrate measurement evidence through data-fidelity updates, model-based reconstruction, sinogram completion, and degradation-aware sampling \citep{liu2023dolce,chung2023diffusionmbir,yang2025ctsdm,chen2025cddm,chen2025cvgdiff}. Their generative states are dense images, volumes, or projections in which global anatomical organization and residual local detail share one parameterization. Joint degradation couples directional information loss with signal-dependent noise, creating distinct recovery demands for long-range structure and local tissue transitions.

Continuous representations provide spatially adaptive parameterization through implicit fields and explicit primitives \citep{yang2026representation}. DPER combines a coordinate-based attenuation field with a diffusion prior for sparse-view and limited-angle CT \citep{du2024dper}. Explicit 3D Gaussians expose position, orientation, scale, and contribution as locally controllable attributes \citep{kerbl20233d}. Recent Gaussian CT methods span projection synthesis, per-volume primitive fitting, learned Gaussian prediction, and discretized Gaussian reconstruction \citep{nikolakakis2024gaspct,li20253dgr,lin2024difgaussian,wu2025dgr}. These studies establish the value of explicit Gaussian representations and generally use one bank for global attenuation geometry and residual detail. The functional hierarchy between these two components consequently remains implicit under joint degradation.

We address this gap with \method{}, a feed-forward hierarchical Gaussian diffusion framework for constrained CT reconstruction across isolated, paired, and joint acquisition settings. Our central premise is that ambiguity from sparse angular sampling, restricted angular coverage, and low photon flux should be decomposed spatially and hierarchically. Spatial decomposition allocates Gaussian capacity to anatomically informative regions. Hierarchical decomposition recovers low-to-mid-frequency attenuation geometry before residual boundaries and tissue transitions.

\method{} realizes this principle through landmark-free anatomical anchors, a foreground capacity field, and sequential structure-to-detail diffusion. The foreground field estimates the utility of allocating Gaussian primitives and modulating their denoising features, allowing the model to emphasize boundaries and residual-sensitive regions. The learned structure bank conditions the detail bank. A shared gradient-isolated network compensates for residual errors under the fixed primitive budget and preserves the rendered Gaussian field as the reconstruction backbone.

The main contributions are summarized below.
\begin{enumerate}[leftmargin=*,itemsep=2pt]
    \item We formulate a unified CT reconstruction task spanning isolated, paired, and joint degradation settings under severely constrained acquisition.
    \item We introduce a sequential structure-to-detail diffusion framework over explicit Gaussian banks. Foreground-aware initialization allocates primitives to anatomically informative regions. The completed structure bank then conditions detail diffusion, and gradient-isolated refinement preserves the Gaussian reconstruction as the output backbone.
    \item Across three distinct CT benchmark datasets, \method{} achieves the best macro-average SSIM on all three datasets and the best macro-average PSNR on both 3D datasets. On LDCT-PD, it improves macro-average PSNR by 5.81 dB and SSIM by 0.113 over the strongest metric-specific baselines.
\end{enumerate}

\begin{figure*}[!t]
    \centering
    \includegraphics[pagebox=cropbox,width=0.72\textwidth]{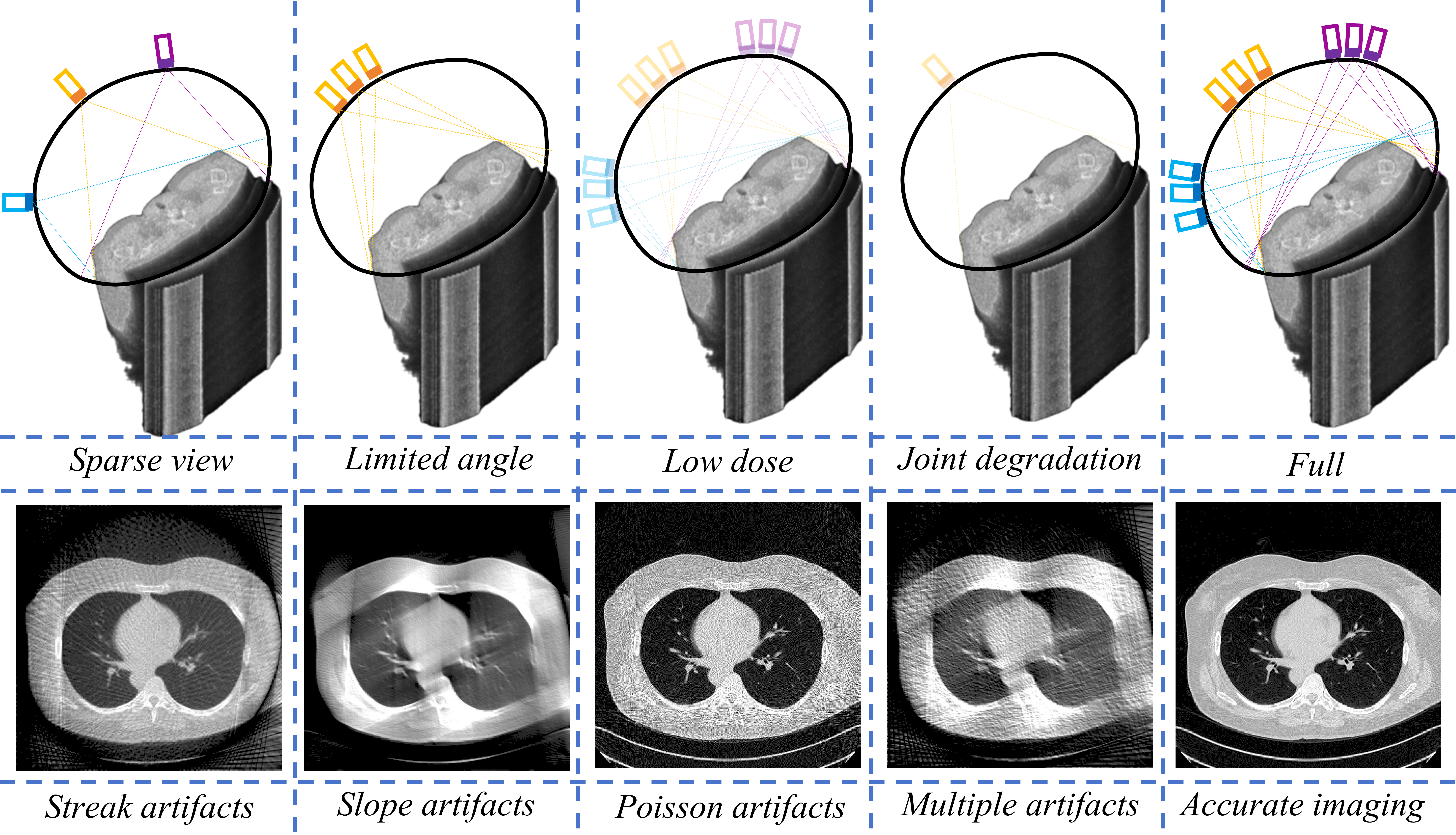}
    \caption{Representative acquisition constraints considered in this study. The figure shows sparse-view sampling, limited-angle coverage, low photon flux, their joint occurrence, and the full reference acquisition at a consistent scale. Paired settings combine the corresponding acquisition factors and are specified in Section~\ref{sec:exp_protocol}.}
    \label{fig:degradations}
\end{figure*}

\section{Related Work}
\label{sec:related}

\subsection{Deep reconstruction under joint CT degradations}

Deep CT reconstruction can be organized by where a learned prior interacts with the imaging operator. Image-domain methods correct an analytic reconstruction, as in RED-CNN and FBPConvNet \citep{chen2017redcnn,jin2017deep}. Frequency-aware methods isolate artifact-related components, as in FreeSeed \citep{ma2023freeseed}. ProCT uses view-aware prompting and contextual examples across incomplete-view settings \citep{ma2024proct}. Unrolled networks such as LEARN insert learned regularizers into iterative reconstruction \citep{chen2018learn}. These paradigms trade direct access to measurements against computational cost and dependence on acquisition-specific training distributions.

Sparse-view, limited-angle, and low-dose acquisition alter the inverse problem in complementary ways, so most learned methods are developed around one dominant factor or a selected pair. DDoCT jointly suppresses projection noise and sparse-view streaks in projection and image domains \citep{li2025ddoct}. Untrained deep-prior reconstruction spans sparse-view, limited-angle, and noisy conditions through per-instance fitting \citep{baguer2022untrained}. When all three degradations occur simultaneously, global non-identifiability and local statistical corruption become coupled. Sparse-view artifact removal can amplify photon noise, and limited-angle reconstruction requires priors that recover missing directional information. \method{} conditions a unified reconstruction path on view density, angular span, and dose metadata.

\subsection{Diffusion models for CT reconstruction}

CT diffusion methods differ chiefly in the variable being diffused and in how measurement consistency is imposed. Image- and volume-domain approaches learn an anatomical score and combine it with an acquisition model. DOLCE uses a transformed-sinogram condition and explicit data-fidelity steps for limited-angle CT \citep{liu2023dolce}, TIFA accelerates limited-angle score sampling through jump, time-reversion, and compressed sampling \citep{wang2024tifa}, and DiffusionMBIR couples pretrained slice priors through a model-based 3D solver \citep{chung2023diffusionmbir}. CDDM cascades latent- and pixel-domain diffusion and mitigates the discrepancy introduced by data-consistency operations \citep{chen2025cddm}. These formulations provide expressive priors over dense image or volume states.

Projection-domain and degradation-process formulations bring the diffusion trajectory closer to CT acquisition. CT-SDM replaces generic Gaussian corruption with progressive projection-view sampling to reconstruct across sampling rates \citep{yang2025ctsdm}. CvG-Diff similarly models sparse-view artifacts as a deterministic degradation process and uses a structure-before-detail sampling strategy \citep{chen2025cvgdiff}. Noise-inspired diffusion adapts the stochastic process to low-dose measurements \citep{gao2025noise}. These studies improve acquisition awareness around one principal degradation axis, with global structure and residual detail represented by the same 3D state.

More generally, multi-stage diffusion can condition a second generative process on the output of a first \citep{singh2024dcdm}. \method{} instantiates this dependency directly in the space of explicit CT Gaussian parameters. The learned structure bank conditions the residual detail bank, establishing a hierarchy between two explicit Gaussian banks within a unified reconstruction path.

\subsection{Continuous and Gaussian representations for tomography}

Continuous representations provide an alternative to dense voxel grids by parameterizing attenuation with coordinates, basis functions, or explicit primitives \citep{yang2026representation}. Implicit neural representations offer continuous coordinate-to-attenuation mappings through repeated ray evaluation and per-instance optimization. DPER couples such a field with a diffusion prior for sparse-view and limited-angle CT through iterative half-quadratic splitting \citep{du2024dper}. Explicit Gaussians expose position, orientation, scale, and contribution as anisotropic primitives \citep{kerbl20233d}. GaSpCT applies Gaussian splatting with foreground and background regularization to CT projection synthesis \citep{nikolakakis2024gaspct}. 3DGR-CT fits FBP-initialized attenuation primitives through differentiable projection \citep{li20253dgr}. DIF-Gaussian predicts Gaussian feature distributions for extremely sparse-view cone-beam CT \citep{lin2024difgaussian}, and DGR reconstructs CT volumes end to end with discretized Gaussian functions \citep{wu2025dgr}. These methods establish the value of continuous Gaussian representations and generally retain a single bank that combines global attenuation geometry with residual detail.

\method{} advances this line by diffusing two foreground-selective Gaussian banks in a structure-conditioned sequence. DiffGS demonstrates generative modeling of Gaussian attributes \citep{zhou2024diffgs}, and UltraGS illustrates how Gaussian attributes and rendering operators can be adapted to medical image formation \citep{yang2026ultrags}. In our CT setting, landmark-free anchors initialize structure primitives, and a continuous foreground field modulates initialization-site selection, attention, and denoiser features \citep{you2022ukpgan,shi2026registers}. The contribution integrates explicit Gaussian representation, spatially selective capacity allocation, and a feed-forward structure-to-detail diffusion hierarchy.

\begin{figure}[!t]
    \centering
    \includegraphics[pagebox=cropbox,width=\linewidth]{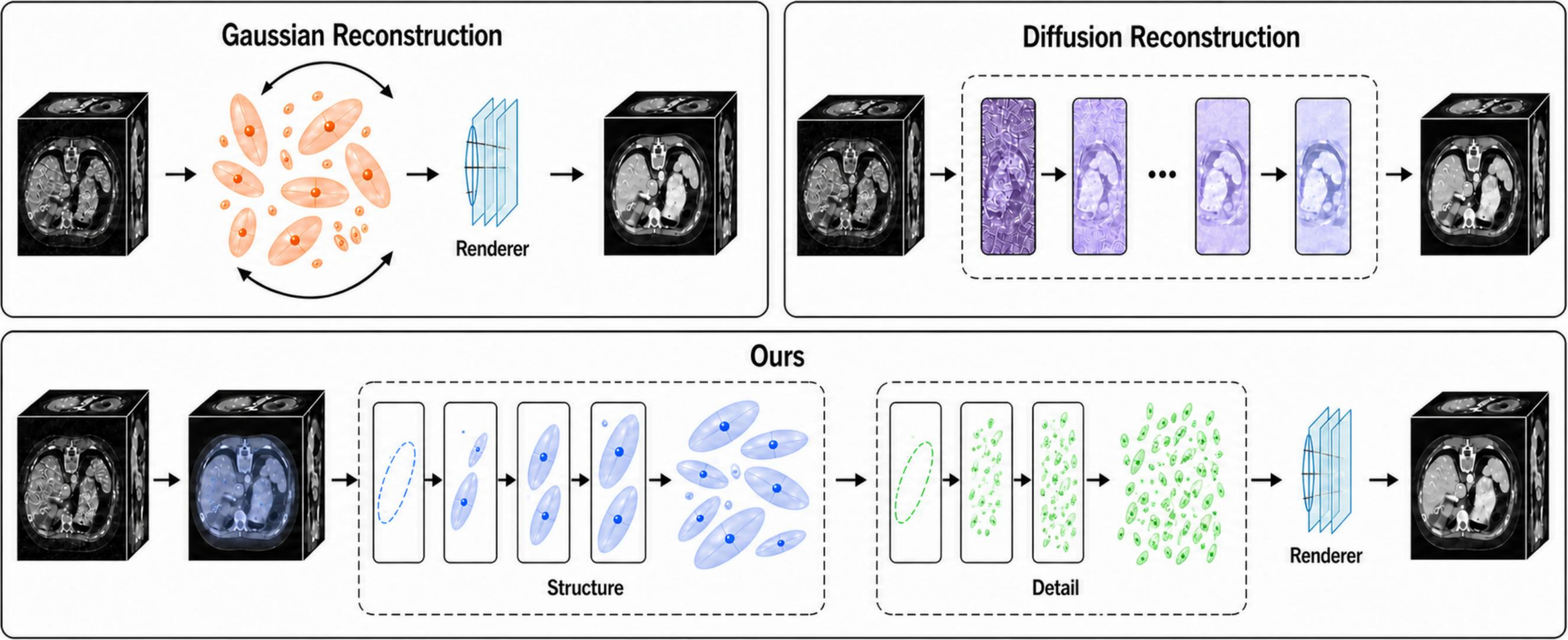}
    \caption{Conceptual comparison of reconstruction paradigms. Conventional Gaussian reconstruction optimizes a single primitive bank. Dense voxel diffusion denoises a single volumetric state. \method{} performs sequential structure and detail diffusion over two explicit Gaussian banks before rendering the reconstructed attenuation field.}
    \label{fig:paradigms}
\end{figure}

Figure~\ref{fig:paradigms} contrasts this design with single-bank Gaussian reconstruction and dense voxel diffusion.

\section{Method}
\label{sec:method}

\begin{figure*}[!t]
    \centering
    \includegraphics[pagebox=cropbox,width=\textwidth]{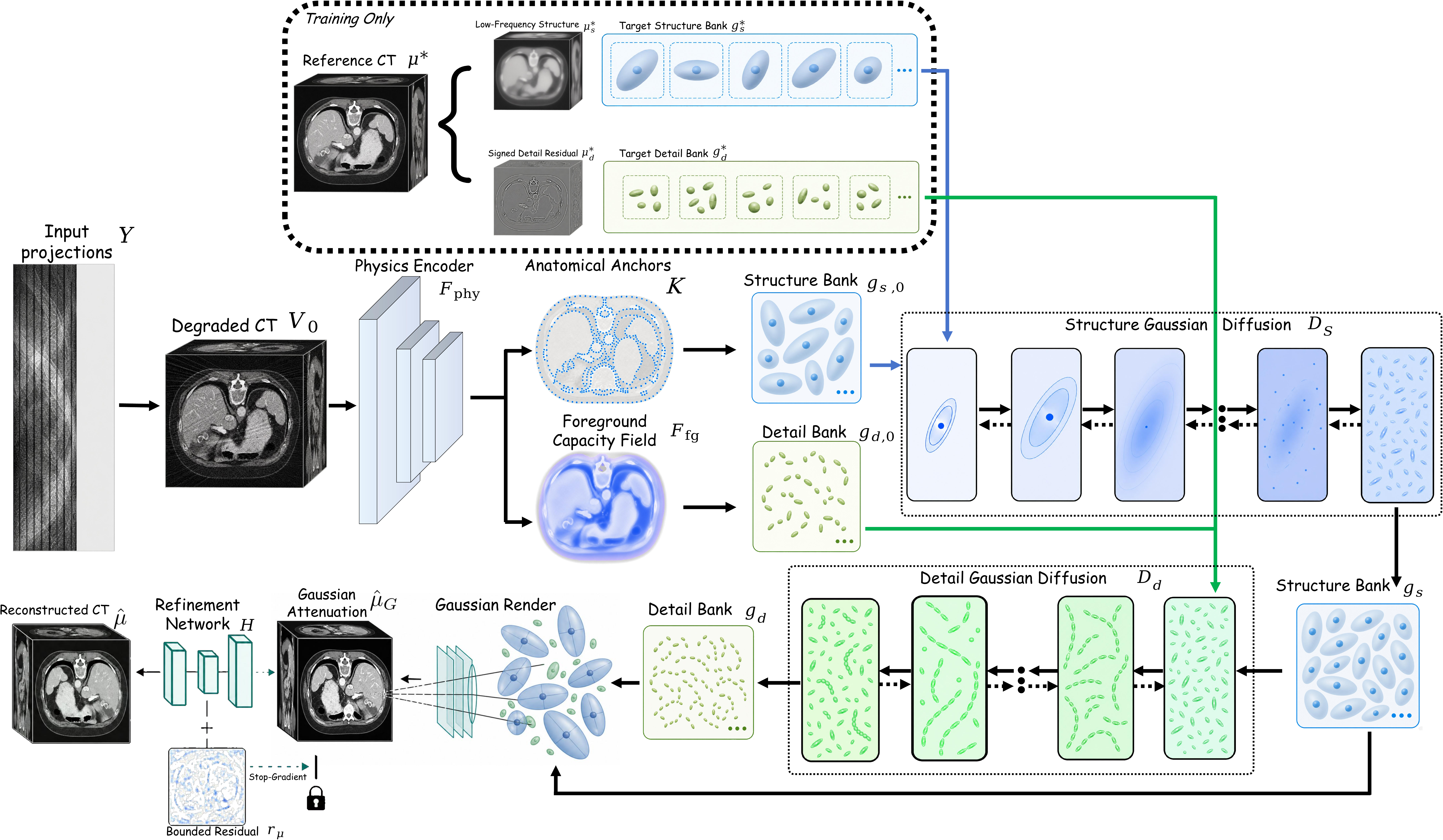}
    \caption{Overview of the proposed \method{} framework. Given jointly degraded measurements $Y$, the physics encoder predicts anatomical anchors $\mathcal K$ and a foreground capacity field $F_{\mathrm{fg}}$ to initialize fixed-cardinality structure and detail Gaussian banks. During training, the reference attenuation field $\mu^{\ast}$ is decomposed to construct slot-aligned Gaussian targets. Structure diffusion $D_s$ first completes global attenuation geometry. Its Gaussian tokens and remaining measurement residual then condition detail diffusion $D_d$. The two completed banks are rendered into the Gaussian attenuation field $\hat\mu_G$, and the gradient-isolated bounded refinement $H$ produces the final reconstruction $\hat\mu$.}
    \label{fig:framework}
\end{figure*}

\paragraph{Overview.}
Given projections affected by sparse-view sampling, limited-angle coverage, low photon counts, or their combinations, \method{} derives physics features and initializes fixed-cardinality structure and detail Gaussian banks from anatomical cues. Slot-aligned residual diffusion first completes the structure bank, whose tokens and unexplained measurement residual then condition detail diffusion. The rendered banks form the reconstruction backbone, followed by a gradient-isolated bounded voxel correction. Figure~\ref{fig:framework} summarizes the pipeline.

\paragraph{Notation.}
The bank index is $b\in\{s,d\}$, with $i$ and $j$ indexing structure and detail primitives. Superscripts $\ast$ and $t$ denote training targets and diffusion time. Subscript $0$, hats, and bars denote initialization, prediction, and exponential-moving-average (EMA) copies. Positions use normalized scanner coordinates in $[-1,1]^3$, and $\odot$ is element-wise multiplication. Stop-gradient preserves its forward value and blocks back-propagation as follows.
$\sg(v)=v$ and $\partial\sg(v)/\partial v=0$.

\subsection{Foreground-Aware Gaussian Initialization}
\label{sec:init}

\method{} initializes both Gaussian banks from a shared physics feature pyramid, landmark-free anatomical anchors, and a continuous foreground capacity field.

\paragraph{Problem formulation.}
Let $\mu^{\ast}$ denote the unknown 3D attenuation field and $Y$ the post-log measurements acquired under geometry $\Theta$.
\begin{equation}
    Y = \project(\mu^{\ast}) + \varepsilon,
    \label{eq:forward}
\end{equation}
where $\project$ is the X-ray transform and $\varepsilon$ is heteroscedastic noise. Let $M_r\in\{0,1\}$ indicate whether ray $r$ is observed and let $I_r=I_0\exp(-Y_r)$ be its estimated photon count. We define the normalized inverse-noise reliability as
\begin{equation}
W_r=M_r\frac{\sqrt{\max(I_r,1)}}
{\max_{\ell:M_\ell=1}\sqrt{\max(I_\ell,1)}}.
\label{eq:reliability}
\end{equation}
Thus, high-count rays receive larger weights and unobserved rays have zero weight in every residual and projection-consistency term. We compute $V_0=\analytic(Y)$ using the geometry-matched FBP operator $\analytic$. The operator $\backproject$ denotes unfiltered backprojection.

\paragraph{Physics feature encoder.}
The multiscale 3D encoder $E_{\mathrm{phy}}$ combines the analytic initialization with the reliability-weighted backprojection of its measurement residual.
\begin{equation}
F_{\mathrm{phy}}=E_{\mathrm{phy}}\!\left(
V_0,\backproject\!\left(
W\odot[Y-\project(V_0)]\right)\right).
\label{eq:physics-features}
\end{equation}
The pyramid $F_{\mathrm{phy}}$ jointly encodes the analytic estimate and its observed-ray inconsistency.
This compact condition is shared by the anchor, foreground, and diffusion modules, so primitive allocation and subsequent denoising are driven by the same acquisition evidence.

\paragraph{Landmark-free anatomical anchors.}
The anchor network $E_{\mathrm{kp}}$ predicts $K$ normalized coordinates, local scales, orientations, confidence values, and structure embeddings.
\begin{equation}
    \mathcal K=E_{\mathrm{kp}}(F_{\mathrm{phy}})
    =\{(c_k,r_k,R_k,\gamma_k,e_k)\}_{k=1}^{K}.
\label{eq:keypoints}
\end{equation}
Here $c_k\in[-1,1]^3$, $r_k\in\R_+^3$, $R_k\in\mathrm{SO}(3)$, $\gamma_k\in[0,1]$, and $e_k\in\R^{d_e}$ denote coordinate, scale, orientation, confidence, and descriptor. Landmark-free training matches the unordered anchors from two geometry-preserving view subsets by a confidence-weighted Hungarian cost over coordinates, orientations, and descriptors. Cross-view consistency and automatically derived edge and distance-transform targets promote repeatability, and repulsion prevents slot collapse.
The resulting anchors form a stable coarse anatomical scaffold under landmark-free supervision, and their geometry determines the initialization of broad structure primitives.

\paragraph{Foreground capacity field.}
The voxel decoder $E_{\mathrm{fg}}$ predicts a scalar capacity field over normalized coordinates $x\in[-1,1]^3$. At diffusion step $t$, this field is sampled at the current centre $m_{b,i}^{t}$ of primitive $i$.
\begin{equation}
\begin{aligned}
F_{\mathrm{fg}}(x)&=E_{\mathrm{fg}}(F_{\mathrm{phy}},\mathcal K)(x)\in[0,1],\\
q_{b,i}^t&=\sg\!\left(F_{\mathrm{fg}}(m_{b,i}^t)\right).
\end{aligned}
\label{eq:foreground}
\end{equation}
The sampled confidence $q_{b,i}^{t}\in[0,1]$ serves as a detached condition for initialization, attention, and denoiser features. $F_{\mathrm{fg}}$ encodes graded primitive utility across regions while the Gaussian attributes determine the diffusion state and attenuation amplitude.
This design concentrates capacity near uncertain boundaries and residual-sensitive regions, stabilizes foreground learning, and preserves the learned attenuation scale during rendering.

\paragraph{Dual Gaussian banks.}
Let $N=N_s+N_d$ be the total primitive budget and $\pi_s\in(0,1)$ the structure fraction. We set $N_s=\operatorname{round}(\pi_sN)$ and $N_d=N-N_s$. The $K$ anchors define the anatomical support for the structure score, while the foreground, edge, and residual responses determine the final fixed-cardinality top-$N_s$ and top-$N_d$ locations. The average structure capacity per anchor is therefore $P_s=N_s/K$; individual anchors may receive different numbers of slots according to their confidence and spatial support. This yields fixed-cardinality banks $\mathcal G_{s,0}$ and $\mathcal G_{d,0}$. With $\operatorname{Samp}(F,m)$ denoting concatenated trilinear samples from pyramid $F$ at $m$, primitive $i$ and its contextual token are
\begin{equation}
\begin{aligned}
g_{b,i}&=(m_{b,i},s_{b,i},u_{b,i},a_{b,i}),\\
z_{b,i}&=\operatorname{Samp}(F_{\mathrm{phy}},m_{b,i}).
\end{aligned}
\label{eq:gaussian-token}
\end{equation}
Here $m_{b,i}\in[-1,1]^3$, $s_{b,i}\in\R_+^3$, and $u_{b,i}\in\R^6$ are centre, principal scales, and a continuous rotation code \citep{zhou2019continuity}. The map $R:\R^6\rightarrow\mathrm{SO}(3)$ returns a 3D rotation, giving $\Sigma_{b,i}=R(u_{b,i})\operatorname{diag}(s_{b,i}^2)R(u_{b,i})^{\top}$. Structure amplitudes satisfy $a_{s,i}\geq0$. Detail amplitudes $a_{d,i}\in\R$ are signed because the residual field $\mu_d^{\ast}=\mu^{\ast}-\mu_s^{\ast}$ contains both positive and negative local corrections. Negative detail amplitudes subtract local overestimation from the structure field, and the final projection $\Pi_+$ in Eq.~\eqref{eq:final-refinement} enforces a nonnegative attenuation output. Diffusion updates $(m,s,u,a)$ and retains $z_{b,i}$ as contextual information. Structure slots inherit anchor geometry, and detail slots use smaller initial scales.
The two banks use the same explicit primitive representation with distinct spatial scales and functional roles. Their fixed cardinalities establish direct slot alignment for Gaussian attributes throughout diffusion.
We denote the complete initialization mapping by $(\mathcal G_{s,0},\mathcal G_{d,0})=\mathcal I(Y,\Theta;\theta_I)$, where $\theta_I$ collects the modules in this subsection.

For any bank $\mathcal G_b$, the renderer $\render$ evaluates a peak-normalized anisotropic Gaussian basis on the voxel coordinate $x$.
\begin{equation}
\begin{aligned}
\mathcal K_G(x;m,\Sigma)
&=\exp\!\left[-\tfrac12(x-m)^{\top}\Sigma^{-1}(x-m)\right],\\
\render(\mathcal G_b)(x)
&=\sum_{i=1}^{N_b}a_{b,i}\mathcal K_G(x;m_{b,i},\Sigma_{b,i}).
\end{aligned}
\label{eq:rendering}
\end{equation}
The renderer converts normalized centres and covariances to physical voxel spacing during evaluation.

\subsection{Structure-to-Detail Hierarchical Gaussian Diffusion}
\label{sec:diffusion}

\method{} organizes Gaussian diffusion as a sequential structure-to-detail process. The structure stage estimates low-to-mid-frequency attenuation geometry, and its completed representation conditions the detail stage that estimates residual boundaries and tissue transitions.

\paragraph{Clean Gaussian targets.}
During training, the reference field is decomposed into
\begin{equation}
    \mu_s^{\ast}=G_{\sigma}*\mu^{\ast},
    \qquad
    \mu_d^{\ast}=\mu^{\ast}-\mu_s^{\ast},
    \label{eq:target-decomposition}
\end{equation}
where $*$ is 3D convolution and $G_{\sigma}$ is a fixed Gaussian low-pass kernel of bandwidth $\sigma$. An EMA copy of the initialization mapping provides slowly varying target partitions.
\begin{equation}
\begin{aligned}
\bar\theta_I&\leftarrow\rho_I\bar\theta_I+(1-\rho_I)\theta_I,\\
(\bar{\mathcal G}_{s,0},\bar{\mathcal G}_{d,0})
&=\mathcal I(Y,\Theta;\bar\theta_I),
\end{aligned}
\label{eq:ema-initialization}
\end{equation}
where $\theta_I$ and $\bar\theta_I$ are the online and EMA parameters, $\rho_I\in[0,1)$ is the decay, and $\leftarrow$ denotes the EMA update. A deterministic operator converts both fields into slot-aligned Gaussian targets.
\begin{equation}
 (\mathcal G_s^{\ast},\mathcal G_d^{\ast})
 =\mathcal T_G(\mu_s^{\ast},\mu_d^{\ast};
 \sg(\bar{\mathcal G}_{s,0}),\sg(\bar{\mathcal G}_{d,0})).
 \label{eq:target-banks}
\end{equation}
The detached EMA banks define Mahalanobis partitions. Weighted moments and local least squares then determine target geometry and amplitude under Eq.~\eqref{eq:rendering}. Detail moments use absolute residual weights and signed amplitude fitting. Empty or near-isotropic slots retain their initial geometry, and principal axes are aligned to the initialization to preserve correspondence.
This construction supplies training targets for every initialized slot. The structure bank approximates the low-pass attenuation field, and the detail bank models its signed residual.
The corresponding fixed-budget oracle field is
\begin{equation}
\mu_G^{\ast}=\render(\mathcal G_s^{\ast})+\render(\mathcal G_d^{\ast}),
\label{eq:oracle-rendering}
\end{equation}
which quantifies the fixed-budget target approximation in evaluation.

\paragraph{Residual diffusion.}
For bank $b$, $\phi_b$ maps Gaussian attributes to a standardized unconstrained vector using inverse hyperbolic tangent for centres, logarithms for positive scales and structure amplitudes, identity for signed detail amplitudes, and the continuous rotation code. The clean slot-aligned correction is
\begin{equation}
    \delta\xi_b^0
    =\phi_b(\mathcal G_b^{\ast})-\phi_b(\mathcal G_{b,0}).
    \label{eq:residual-state}
\end{equation}
Residual diffusion around the physics-derived initialization preserves its spatial layout and concentrates the generative process on acquisition-dependent errors in geometry and attenuation. We use the standard noise-prediction formulation of denoising diffusion probabilistic models \citep{ho2020denoising}.
For a bank-specific schedule of $T_b$ steps, let $\beta_{b,t}$ be the noise variance, $\alpha_{b,t}=1-\beta_{b,t}$, and $\bar\alpha_{b,t}=\prod_{\ell=1}^{t}\alpha_{b,\ell}$. Forward diffusion is
\begin{equation}
\delta\xi_{b,i}^{t}
=\sqrt{\bar\alpha_{b,t}}\,\delta\xi_{b,i}^{0}
+\sqrt{1-\bar\alpha_{b,t}}\,\epsilon_{b,i},
\quad \epsilon_{b,i}\sim\N(0,I),
\label{eq:forward-diffusion}
\end{equation}
where $I$ is the identity covariance and $\bar\alpha_{b,0}=1$. Starting from $\delta\xi_b^{T_b}\sim\N(0,I)$, deterministic DDIM ($\eta=0$) \citep{song2021ddim} uses predicted noise $\widehat\epsilon_{b,i}^{\,t}$ to update
\begin{equation}
\begin{aligned}
\widehat{\delta\xi}_{b,i}^{\,0}
&=\frac{\delta\xi_{b,i}^{t}
-\sqrt{1-\bar\alpha_{b,t}}\,\widehat\epsilon_{b,i}^{\,t}}
{\sqrt{\bar\alpha_{b,t}}},\\
\delta\xi_{b,i}^{\,t-1}
&=\sqrt{\bar\alpha_{b,t-1}}\,
\widehat{\delta\xi}_{b,i}^{\,0}
+\sqrt{1-\bar\alpha_{b,t-1}}\,
\widehat\epsilon_{b,i}^{\,t}.
\end{aligned}
\label{eq:ddim-update}
\end{equation}
Inverse standardization and $\phi_b^{-1}$ add the correction to $\mathcal G_{b,0}$ and recover valid attributes. Centre clipping and positive numerical floors maintain valid domains. The same blockwise statistics are used at training and inference.

\paragraph{Structure Gaussian diffusion.}
At reverse step $t$, each structure primitive samples the physics pyramid at its current centre and fuses this evidence with its initial token and Gaussian attributes.
\begin{equation}
\begin{aligned}
f_{s,i}^{t}&=\operatorname{Samp}(F_{\mathrm{phy}},m_{s,i}^{t}),\\
c_{s,i}^{t}&=w(q_{s,i}^{t})M_s\!\left(
[z_{s,i},f_{s,i}^{t},\phi_s(g_{s,i,0})]\right),\\
\widehat\epsilon_{s,i}^{\,t}
&=D_s(\delta\xi_{s,i}^{t},c_{s,i}^{t},t).
\end{aligned}
\label{eq:structure-diffusion}
\end{equation}
Here $[\cdot]$ is concatenation, $M_s$ is a fusion MLP, and $w(q)=0.15+0.85q$. The permutation-equivariant set denoiser $D_s$ returns a noise vector matching $\delta\xi_{s,i}^{t}$ and adds $\log(10^{-6}+q_{s,i}^{t})$ to its self-attention logits. Repeating Eq.~\eqref{eq:ddim-update} yields $\mathcal G_s$, which is rendered and encoded as
\begin{equation}
\begin{aligned}
    \hat\mu_s&=\render(\mathcal G_s),\\
    R_s=\{r_{s,i}\}_{i=1}^{N_s}
    &=E_s\!\left(\mathcal G_s,\hat\mu_s,F_{\mathrm{phy}}\right).
\end{aligned}
\label{eq:structure-prior}
\end{equation}
$E_s$ returns one token $r_{s,i}$ per completed structure primitive.
These tokens summarize the recovered long-range attenuation organization and form the explicit bridge from the first diffusion stage to the second.

\paragraph{Structure-conditioned detail Gaussian diffusion.}
The detail stage receives $R_s$ and the weighted backprojection of the remaining measurement residual,
\begin{equation}
    B_s=\backproject\!\left(
    W\odot\left[Y-\project(\hat\mu_s)\right]\right).
\label{eq:structure-residual}
\end{equation}
Here $B_s$ is the unexplained observed-ray evidence in the voxel domain. Training constructs detail conditions from a completed structure sample. Let $\theta_s$ and $\bar\theta_s$ parameterize $D_s$ and its EMA copy $\bar D_s$, updated with decay $\rho_s$ as $\bar\theta_s\leftarrow\rho_s\bar\theta_s+(1-\rho_s)\theta_s$. The operator $\operatorname{DDIM}_{T_s}$ denotes $T_s$ applications of Eq.~\eqref{eq:ddim-update}. Each detail update constructs detached structure conditions as
\begin{equation}
\begin{aligned}
\widetilde{\mathcal G}_s
&=\operatorname{DDIM}_{T_s}
  (\mathcal G_{s,0};F_{\mathrm{phy}},F_{\mathrm{fg}},\bar D_s),\\
\widetilde\mu_s&=\render(\widetilde{\mathcal G}_s),\\
\check R_s
&=\sg\!\left(E_s(\widetilde{\mathcal G}_s,
\widetilde\mu_s,F_{\mathrm{phy}})\right),\\
\check B_s
&=\sg\!\left(\backproject\!\left(
W\odot[Y-\project(\widetilde\mu_s)]\right)\right).
\end{aligned}
\label{eq:training-structure-condition}
\end{equation}
Tildes denote the completed EMA prediction, and checks denote stopped training conditions. $(\check R_s,\check B_s)$ replaces $(R_s,B_s)$ during training. Each detail primitive then samples the physics and residual volumes and aggregates its $k$ nearest structure tokens.
\begin{equation}
\begin{aligned}
f_{d,j}^{t}&=\operatorname{Samp}(F_{\mathrm{phy}},m_{d,j}^{t}),\\
b_{d,j}^{t}&=\operatorname{Samp}(B_s,m_{d,j}^{t}),\\
h_{d,j}^{t}&=\operatorname{LocalAttn}\!\left(
z_{d,j},\{r_{s,i}\}_{i\in\mathcal N_k(m_{d,j}^{t})}\right),\\
c_{d,j}^{t}&=w(q_{d,j}^{t})M_d\!\left(
[z_{d,j},\phi_d(g_{d,j,0}),f_{d,j}^{t},b_{d,j}^{t},h_{d,j}^{t}]\right),\\
\widehat\epsilon_{d,j}^{\,t}
&=D_d(\delta\xi_{d,j}^{t},c_{d,j}^{t},t).
\end{aligned}
    \label{eq:detail-diffusion}
\end{equation}
Here $\mathcal N_k(m)$ contains the $k$ nearest structure centres. Restricted scaled dot-product attention \citep{vaswani2017attention} is $h_{d,j}^t=\sum_i\alpha_{ji}Vr_{s,i}$, where $\alpha_{ji}=\operatorname{softmax}_i((Qz_{d,j})^\top Kr_{s,i}/\sqrt{d_h})$ and $Q,K,V$ are learned projections of key dimension $d_h$. The fusion MLP $M_d$ combines local physics, unexplained measurements, and structure context. $D_d$ predicts noise with the dimension of $\delta\xi_{d,j}^t$. This explicit dependency implements sequential structure-to-detail diffusion.
Local structure attention limits detail updates to anatomically compatible corrections, and $B_s$ directs them toward the remaining measurement residual.

\subsection{Gaussian-Conditioned Attenuation Refinement}
\label{sec:refinement}

The completed banks form the additive Gaussian reconstruction
\begin{equation}
\hat\mu_d=\render(\mathcal G_d),
\qquad
\hat\mu_G=\hat\mu_s+\hat\mu_d.
\label{eq:rendered-fields}
\end{equation}
$\hat\mu_G$ is evaluated in physical voxel spacing, with $a_{b,i}$ in attenuation units. Foreground confidence influences placement and denoising. Rendering uses the learned attenuation amplitudes directly. The renderer expands a finite set of smooth anisotropic 3D basis functions onto a dense 3D voxel grid. This representation captures global attenuation organization efficiently, while finite primitive capacity leaves grid-scale errors at sharp interfaces, narrow structures, and regions affected by structured acquisition artifacts. The refinement network $H$ predicts the corresponding bounded voxel-domain residual.
\begin{equation}
\begin{aligned}
    B_G&=\backproject\!\left(
      W\odot[Y-\project(\sg(\hat\mu_G))]\right),\\
    r_{\mu}&=r_{\max}\tanh H\!\left(
    \operatorname{cat}[\sg(\hat\mu_G),\sg(B_G),F_{\mathrm{fg}}]\right), \\
    \hat\mu&=\Pi_+\!\left(\sg(\hat\mu_G)+r_{\mu}\right),\\
    \Pi_+(v)&=\max(v,0).
\end{aligned}
\label{eq:final-refinement}
\end{equation}
Here $H$ is a multiscale 3D residual network operating on the rendered attenuation volume, the observed-ray residual backprojection, and the foreground field. $\operatorname{cat}$ is channel-wise concatenation, and $r_{\max}$ bounds the correction by $\|r_\mu\|_{\infty}\leq r_{\max}$. The single nonnegative projection $\Pi_+$ follows residual addition. The Gaussian hierarchy supplies the global 3D attenuation layout and localized tissue representation, while $H$ corrects residual errors after dense voxelization. Stop-gradient and Stage-II freezing maintain this division during refinement training.

\subsection{Training Objectives and Feed-Forward Inference}
\label{sec:training}

Training has two stages. Stage I alternates structure updates, which learn the initialization and $D_s$, with detail updates, which construct the stopped conditions in Eq.~\eqref{eq:training-structure-condition} and learn $D_d$. Stage II trains the refinement network after freezing all Stage-I modules.

\paragraph{Stage I Gaussian representation learning.}
All $\lambda$ terms are nonnegative weights. We group anchor consistency, structural reconstruction, and slot repulsion in $\mathcal L_{\mathrm{kp}}$, and capacity-target cross-entropy and smoothness in $\mathcal L_{\mathrm{fg}}$, giving $\mathcal L_{\mathrm{aux}}=\lambda_{\mathrm{kp}}\mathcal L_{\mathrm{kp}}+\lambda_{\mathrm{fg}}\mathcal L_{\mathrm{fg}}$. The diffusion loss is
\begin{equation}
\mathcal L_{\mathrm{diff}}
=\sum_{b\in\{s,d\}}\lambda_b\,
\mathbb E_{t,i,\epsilon}
\!\left[w(q_{b,i}^t)
\left\|\epsilon_{b,i}-\widehat\epsilon_{b,i}^{\,t}\right\|_2^2\right].
\label{eq:diffusion-losses}
\end{equation}
The expectation samples a timestep, primitive, and noise for each bank. Target and scale separation is enforced by
\begin{equation}
\begin{aligned}
\mathcal L_{\mathrm{sep}}={}&
\lambda_{\mathrm{str}}\|\hat\mu_s-\mu_s^{\ast}\|_1
+\lambda_{\mathrm{det}}\|\hat\mu_d-\mu_d^{\ast}\|_1\\
&+\lambda_{\mathrm{bw}}\!\left(
\sum_i[\tau_s-\overline s_{s,i}]_+
+\sum_j[\overline s_{d,j}-\tau_d]_+\right),
\end{aligned}
\label{eq:bank-losses}
\end{equation}
where $\overline s_{b,i}$ is the mean principal scale, $[x]_+=\max(x,0)$, and $\tau_s,\tau_d$ encourage broad structure and compact detail primitives. For any attenuation field $\nu$, observed-ray consistency is
\begin{equation}
\mathcal L_{\mathrm{proj}}(\nu)
=\left\|W\odot\left(\project(\nu)-Y\right)\right\|_1.
\label{eq:projection-loss}
\end{equation}
Together, the bank-specific targets and scale penalty establish distinct structure and detail roles through supervision and architecture.
The complete Stage-I objective is
\begin{equation}
\mathcal L_G=
\mathcal L_{\mathrm{diff}}+\mathcal L_{\mathrm{aux}}
+\mathcal L_{\mathrm{sep}}
+\lambda_{\mathrm{joint}}\|\hat\mu_G-\mu^{\ast}\|_1
+\lambda_{\mathrm{proj}}\mathcal L_{\mathrm{proj}}(\hat\mu_G).
\label{eq:gaussian-loss}
\end{equation}

\paragraph{Stage II attenuation refinement.}
After freezing all Stage-I modules, we optimize $H$ with
\begin{equation}
\begin{aligned}
\mathcal L_H={}&
\|\hat\mu-\mu^{\ast}\|_1
+\lambda_{\mathrm{ssim}}
\bigl(1-\operatorname{SSIM}_{3\mathrm D}(\hat\mu,\mu^{\ast})\bigr)\\
&+\lambda_{\mathrm{proj}}\mathcal L_{\mathrm{proj}}(\hat\mu)
+\lambda_{\mathrm{res}}\|r_{\mu}\|_1.
\end{aligned}
\label{eq:refinement-loss}
\end{equation}
Here $\operatorname{SSIM}_{3\mathrm D}$ is volumetric structural similarity, and the last term keeps the bounded correction small.

\paragraph{Feed-forward inference.}
At test time, \method{} applies $T_s$ structure and $T_d$ detail DDIM steps, renders both banks, and performs one bounded refinement pass.

\section{Experiments}
\label{sec:experiments}
\subsection{Experimental setup}
\label{sec:exp_protocol}

We evaluate \method{} on the chest subset of the public Low Dose CT Image and Projection Data collection (LDCT-and-Projection-data; hereafter LDCT-PD) \citep{mccollough2020ldct}, the public TCIA Pancreas collection \citep{roth2016pancreasct,roth2015deeporgan,clark2013tcia}, and the slice-based LoDoPaB-CT benchmark \citep{leuschner2019lodopabdataset,leuschner2021lodopab}. Table~\ref{tab:data_protocol} reports the anatomical region, reconstruction dimensionality, and the number of training or fine-tuning and test samples. The 3D model is trained on 11 LDCT-PD volumes and tested on 45 LDCT-PD volumes. TCIA Pancreas is used for cross-dataset generalization, and a separate 2D model is trained on 3,007 LoDoPaB-CT slices and tested on 12,830 slices. All partitions are patient-disjoint; all slices from one LoDoPaB-CT subject remain in the same partition.

\begin{table}[t]
    \centering
    \caption{Anatomical region, reconstruction dimensionality, and data split for each dataset.}
    \label{tab:data_protocol}
    \small
    \setlength{\tabcolsep}{5pt}
    \resizebox{\columnwidth}{!}{%
    \begin{tabular}{@{}lcccc@{}}
        \toprule
        Dataset & Region & Dim. & Train/Fine-tune & Test \\
        \midrule
        LDCT-PD & Thorax & 3D & 11 volumes & 45 volumes \\
        TCIA Pancreas & Abdomen & 3D & 5 volumes & 36 volumes \\
        LoDoPaB-CT & Thorax & 2D & 3,007 slices & 12,830 slices \\
        \bottomrule
    \end{tabular}}
\end{table}

The 16 fixed acquisition settings form the four groups used in Tables~\ref{tab:diffct_results}--\ref{tab:tcia_results}. Limited-angle acquisition uses angular spans $0^{\circ}$--$\theta$ with $\theta\in\{90^{\circ},\allowbreak120^{\circ},\allowbreak150^{\circ},\allowbreak180^{\circ}\}$. Sparse-view acquisition retains view ratios $r\in\{1/3,\allowbreak1/6,\allowbreak1/9,\allowbreak1/12\}$ over the nominal angular range. Their paired combination uses $(r,\theta)\in\{(1/3,90^{\circ}),\allowbreak(1/6,120^{\circ}),\allowbreak(1/9,150^{\circ}),\allowbreak(1/12,180^{\circ})\}$. The final group contains standalone low dose and low dose combined with $(1/3,120^{\circ})$, $(1/6,150^{\circ})$, or $(1/9,180^{\circ})$. Standalone low dose uses the paired low-dose observation for datasets that provide it. Derived joint settings use deterministic Poisson projection corruption. The result tables label each setting by $\theta$, $r$, or $(r,\theta)$.

\paragraph{Implementation and hyperparameters.}
The main configuration uses $K=128$ anatomical anchors and a total budget of $N=512$ Gaussian primitives. The structure fraction is $\pi_s=0.40$, giving $N_s=205$ structure primitives, $N_d=307$ detail primitives, and an average $P_s=N_s/K\approx1.60$ structure slots per anchor. The target decomposition uses $\sigma=1$ voxel, each detail primitive attends to the $k=8$ nearest structure primitives, and inference uses $T_s=T_d=3$ deterministic DDIM steps. The refinement network has 24 hidden channels and uses $r_{\max}=0.30$ in the normalized attenuation range. These values are selected on the validation split and fixed across datasets and degradation settings. Training uses AdamW \citep{loshchilov2019decoupled} with weight decay $10^{-4}$. Experiments are run in the same PyTorch/CUDA environment on a single NVIDIA RTX PRO 6000 GPU.

For each acquisition setting on TCIA Pancreas, all methods start from the corresponding LDCT-PD-trained checkpoint and are fine-tuned on the same five target-domain volumes for five epochs with AdamW and a learning rate of $10^{-5}$. Evaluation uses the same 36 disjoint target-domain volumes. Preprocessing, target normalization, checkpoint selection, and data partitions are shared across methods.

The comparison includes the Gaussian-based 3DGR-CT \citep{li20253dgr}, the diffusion methods DOLCE \citep{liu2023dolce}, NEED \citep{gao2025noise}, and TIFA \citep{wang2024tifa}, and the deterministic methods FBPConvNet \citep{jin2017deep}, FreeSeed \citep{ma2023freeseed}, ProCT \citep{ma2024proct}, and U-Net \citep{ronneberger2015u}. DOLCE, FreeSeed, ProCT, and TIFA use geometry-matched 3D implementations on LDCT-PD. NEED and 3DGR-CT use project-local hybrid implementations, and the LoDoPaB-CT comparison uses 2D configurations.

We report peak signal-to-noise ratio (PSNR, dB) and structural similarity index measure (SSIM) \citep{wang2004image} on attenuation fields clipped and normalized to $[0,1]$. PSNR uses a data range of 1. SSIM uses $C_1=0.01^2$ and $C_2=0.03^2$ and is computed over each complete 3D volume for LDCT-PD and TCIA Pancreas and over each 2D slice for LoDoPaB-CT. Metrics are then averaged over all held-out volumes or slices for each setting. Each dataset table contains all 16 setting-level results and their unweighted macro-average. The best and second-best results are indicated by boldface and underlining.

\subsection{Overall quantitative comparison}

Across the two volumetric benchmarks, \method{} ranks first in both metrics for all 16 acquisition settings. Its advantage remains pronounced in the three rows where sparse-view, limited-angle, and low-dose degradation occur simultaneously. On the 2D LoDoPaB-CT benchmark, \method{} is effectively tied with DOLCE in macro-average PSNR and gives the highest mean SSIM and the most consistent performance under the joint settings. The following three subsections analyze the result tables separately.

\begin{table*}[!t]
\centering
\begin{minipage}{\textwidth}
    \centering
    \captionof{table}{Results on LDCT-PD (3D). Each entry reports PSNR in dB followed by SSIM. The four setting groups are defined in Section~\ref{sec:exp_protocol}. Best and second-best values are shown in bold and underlined.}
    \label{tab:diffct_results}
    \scriptsize
    \setlength{\tabcolsep}{2.2pt}
    \renewcommand{\arraystretch}{0.98}
    \resizebox{\textwidth}{!}{%
    \begin{tabular}{@{}lccccccccc@{}}
        \toprule
        Setting & \shortstack{FBPConvNet\\TIP'17} & \shortstack{U-Net\\MICCAI'15} & \shortstack{FreeSeed\\MICCAI'23} & \shortstack{ProCT\\PR'26} & \shortstack{DOLCE\\ICCV'23} & \shortstack{NEED\\MIA'25} & \shortstack{TIFA\\TMI'24} & \shortstack{3DGR-CT\\MIA'25} & \shortstack{\method{}\\Ours} \\
        \midrule
        \multicolumn{10}{@{}l}{\textit{Limited-angle}} \\
        $90^\circ$ & 18.928/0.5733 & 20.271/\underline{0.6023} & 18.664/0.5804 & \underline{20.597}/0.5500 & 17.915/0.5583 & 20.263/0.6018 & 18.828/0.5867 & 20.278/0.6004 & \textbf{25.306}/\textbf{0.7959} \\
        $120^\circ$ & 22.557/0.6868 & 23.967/\underline{0.7221} & 22.277/0.6388 & \underline{24.020}/0.6241 & 21.489/0.6300 & 23.887/0.7172 & 22.362/0.6360 & 23.913/0.7178 & \textbf{29.172}/\textbf{0.8598} \\
        $150^\circ$ & 26.142/0.7506 & 27.101/\underline{0.7734} & 25.174/0.6543 & 26.489/0.6463 & 24.852/0.6256 & 27.032/0.7722 & 25.726/0.6705 & \underline{27.131}/0.7717 & \textbf{33.097}/\textbf{0.9046} \\
        $180^\circ$ & 30.055/0.8044 & \underline{30.180}/\underline{0.8121} & 26.111/0.6753 & 28.414/0.6385 & 27.516/0.6436 & 29.980/0.8098 & 29.015/0.7051 & 30.016/0.8115 & \textbf{41.247}/\textbf{0.9687} \\
        \midrule
        \multicolumn{10}{@{}l}{\textit{Sparse-view}} \\
        $1/3$ & 29.869/0.7971 & \underline{30.183}/\underline{0.8119} & 26.081/0.6760 & 28.404/0.6382 & 27.507/0.6436 & 29.903/0.8095 & 29.011/0.7036 & 30.037/0.8114 & \textbf{40.357}/\textbf{0.9519} \\
        $1/6$ & 29.560/0.7889 & \underline{30.186}/0.8126 & 25.997/0.6794 & 28.390/0.6393 & 27.459/0.6447 & 29.921/0.8091 & 28.958/0.6985 & 30.155/\underline{0.8131} & \textbf{37.922}/\textbf{0.9259} \\
        $1/9$ & 28.520/0.7389 & \underline{30.097}/0.8115 & 25.955/0.6734 & 28.320/0.6407 & 27.313/0.6416 & 29.850/0.8078 & 28.823/0.6890 & 29.997/\underline{0.8116} & \textbf{36.313}/\textbf{0.9043} \\
        $1/12$ & 27.250/0.6698 & \underline{29.787}/0.8037 & 25.840/0.6567 & 28.176/0.6393 & 27.058/0.6319 & 29.583/0.8022 & 28.558/0.6700 & 29.677/\underline{0.8039} & \textbf{35.044}/\textbf{0.8834} \\
        \midrule
        \multicolumn{10}{@{}l}{\textit{Sparse-view $+$ limited-angle}} \\
        $(1/3,90^\circ)$ & 18.891/0.5673 & 20.232/\underline{0.6022} & 18.607/0.5768 & \underline{20.569}/0.5491 & 17.887/0.5557 & 20.232/0.6016 & 18.792/0.5840 & 20.242/0.5995 & \textbf{25.132}/\textbf{0.7855} \\
        $(1/6,120^\circ)$ & 22.262/0.6517 & 23.957/\underline{0.7204} & 22.231/0.6380 & \underline{23.974}/0.6227 & 21.433/0.6254 & 23.875/0.7165 & 22.323/0.6305 & 23.900/0.7150 & \textbf{28.663}/\textbf{0.8291} \\
        $(1/9,150^\circ)$ & 25.268/0.6708 & \underline{27.057}/\underline{0.7700} & 25.087/0.6554 & 26.377/0.6413 & 24.660/0.6172 & 27.029/0.7699 & 25.592/0.6573 & 27.049/0.7690 & \textbf{31.698}/\textbf{0.8547} \\
        $(1/12,180^\circ)$ & 27.180/0.6688 & \underline{29.722}/\underline{0.8038} & 25.840/0.6567 & 28.176/0.6393 & 27.054/0.6318 & 29.527/0.8011 & 28.578/0.6700 & 29.616/0.8034 & \textbf{35.028}/\textbf{0.8853} \\
        \midrule
        \multicolumn{10}{@{}l}{\textit{Low-dose and joint degradation}} \\
        LD & 28.651/0.7421 & \underline{29.506}/0.8094 & 25.789/0.6807 & 28.245/0.6327 & 27.241/0.6309 & 29.464/0.8091 & 28.718/0.6591 & 29.492/\underline{0.8116} & \textbf{36.292}/\textbf{0.9039} \\
        \cmidrule(lr){1-10}
        $(1/3,120^\circ)$ & 19.436/0.3626 & \underline{23.475}/\underline{0.6962} & 20.759/0.4406 & 23.086/0.5028 & 20.240/0.4355 & 23.453/0.6939 & 21.319/0.4449 & 23.456/0.6919 & \textbf{27.172}/\textbf{0.7781} \\
        $(1/6,150^\circ)$ & 19.427/0.2829 & \underline{26.012}/\underline{0.7375} & 21.800/0.3881 & 24.711/0.4959 & 21.755/0.4139 & 25.985/0.7360 & 23.434/0.4689 & 25.994/0.7343 & \textbf{29.152}/\textbf{0.8032} \\
        $(1/9,180^\circ)$ & 19.116/0.2336 & 27.808/0.7582 & 21.458/0.3200 & 25.702/0.4810 & 22.170/0.3969 & \underline{27.861}/\underline{0.7584} & 24.346/0.4702 & 27.767/0.7578 & \textbf{30.878}/\textbf{0.8228} \\
        \midrule
        Mean & 24.569/0.6243 & \underline{26.846}/\underline{0.7530} & 23.604/0.5994 & 25.853/0.5988 & 23.972/0.5829 & 26.740/0.7510 & 25.274/0.6215 & 26.795/0.7515 & \textbf{32.655}/\textbf{0.8661} \\
        \bottomrule
    \end{tabular}}

\vspace{1.2em}

    \centering
    \captionof{table}{Results on LoDoPaB-CT (2D). Each entry reports PSNR in dB followed by SSIM. The four setting groups are defined in Section~\ref{sec:exp_protocol}. Best and second-best values are shown in bold and underlined.}
    \label{tab:lodopab_results}
    \scriptsize
    \setlength{\tabcolsep}{2.2pt}
    \renewcommand{\arraystretch}{0.98}
    \resizebox{\textwidth}{!}{%
    \begin{tabular}{@{}lccccccccc@{}}
        \toprule
        Setting & \shortstack{FBPConvNet\\TIP'17} & \shortstack{U-Net\\MICCAI'15} & \shortstack{FreeSeed\\MICCAI'23} & \shortstack{ProCT\\PR'26} & \shortstack{DOLCE\\ICCV'23} & \shortstack{NEED\\MIA'25} & \shortstack{TIFA\\TMI'24} & \shortstack{3DGR-CT\\MIA'25} & \shortstack{\method{}\\Ours} \\
        \midrule
        \multicolumn{10}{@{}l}{\textit{Limited-angle}} \\
        $90^\circ$ & 29.459/0.8085 & 27.618/0.7275 & 29.435/0.8191 & 29.630/0.8294 & \textbf{31.577}/\underline{0.8654} & 30.576/0.8558 & 31.006/0.8606 & 27.801/0.7330 & \underline{31.434}/\textbf{0.8778} \\
        $120^\circ$ & 31.933/0.8669 & 29.032/0.7570 & 31.361/0.8475 & 31.813/0.8727 & \underline{34.151}/\underline{0.9065} & 32.905/0.8960 & 33.484/0.9002 & 29.451/0.7664 & \textbf{34.215}/\textbf{0.9170} \\
        $150^\circ$ & 35.164/0.9178 & 30.591/0.7951 & 33.881/0.8955 & 35.011/0.9224 & \textbf{37.595}/\underline{0.9427} & 36.228/0.9339 & 36.866/0.9398 & 31.189/0.8038 & \underline{37.509}/\textbf{0.9507} \\
        $180^\circ$ & 49.962/0.9902 & 35.130/0.8653 & 48.122/0.9835 & 49.731/0.9898 & 50.488/0.9902 & \textbf{51.106}/\textbf{0.9911} & \underline{50.943}/\underline{0.9910} & 36.897/0.9068 & 49.982/0.9906 \\
        \midrule
        \multicolumn{10}{@{}l}{\textit{Sparse-view}} \\
        $1/3$ & 47.408/0.9823 & 35.006/0.8630 & 46.191/0.9749 & 47.341/0.9818 & 48.268/0.9831 & \textbf{48.512}/\textbf{0.9837} & \underline{48.349}/0.9835 & 37.029/0.9068 & 47.801/\underline{0.9836} \\
        $1/6$ & 43.544/0.9620 & 34.630/0.8467 & 42.586/0.9540 & 43.362/0.9616 & \textbf{44.280}/0.9649 & 44.142/0.9644 & \underline{44.201}/\underline{0.9651} & 36.514/0.8958 & 44.164/\textbf{0.9667} \\
        $1/9$ & 41.235/0.9437 & 33.874/0.8280 & 40.037/0.9240 & 41.085/0.9438 & \underline{41.882}/\underline{0.9487} & 41.778/0.9480 & 41.779/0.9486 & 35.767/0.8834 & \textbf{41.908}/\textbf{0.9521} \\
        $1/12$ & 39.656/0.9282 & 33.485/0.8156 & 38.533/0.9050 & 39.519/0.9281 & \underline{40.313}/\underline{0.9351} & 40.185/0.9344 & 40.156/0.9342 & 35.431/0.8745 & \textbf{40.406}/\textbf{0.9397} \\
        \midrule
        \multicolumn{10}{@{}l}{\textit{Sparse-view $+$ limited-angle}} \\
        $(1/3,90^\circ)$ & 29.496/0.8070 & 27.603/0.7270 & 29.417/0.8171 & 29.661/0.8286 & \textbf{31.534}/\underline{0.8623} & 30.590/0.8532 & 30.992/0.8574 & 27.930/0.7361 & \underline{31.437}/\textbf{0.8743} \\
        $(1/6,120^\circ)$ & 31.695/0.8540 & 28.932/0.7516 & 31.121/0.8285 & 31.615/0.8566 & \underline{33.768}/\underline{0.8872} & 32.720/0.8777 & 33.107/0.8796 & 29.351/0.7594 & \textbf{33.942}/\textbf{0.9003} \\
        $(1/9,150^\circ)$ & 34.209/0.8852 & 30.149/0.7728 & 33.038/0.8544 & 34.065/0.8836 & \underline{36.371}/\underline{0.9081} & 35.290/0.9000 & 35.596/0.9008 & 30.858/0.7835 & \textbf{36.415}/\textbf{0.9192} \\
        $(1/12,180^\circ)$ & 39.655/0.9281 & 33.208/0.8122 & 38.536/0.9050 & 39.518/0.9281 & \underline{40.314}/\underline{0.9352} & 40.184/0.9343 & 40.159/0.9343 & 35.440/0.8752 & \textbf{40.433}/\textbf{0.9397} \\
        \midrule
        \multicolumn{10}{@{}l}{\textit{Low-dose and joint degradation}} \\
        LD & 39.414/0.9244 & 34.435/0.8408 & 39.035/0.9206 & 39.341/0.9265 & \underline{39.904}/\underline{0.9309} & 39.795/0.9305 & 39.771/0.9304 & 35.575/0.8836 & \textbf{39.999}/\textbf{0.9342} \\
        \cmidrule(lr){1-10}
        $(1/3,120^\circ)$ & 31.337/0.8203 & 28.700/0.7250 & 30.541/0.7880 & 30.799/0.8207 & \underline{32.752}/\underline{0.8516} & 31.754/0.8385 & 32.115/0.8413 & 29.374/0.7441 & \textbf{32.895}/\textbf{0.8633} \\
        $(1/6,150^\circ)$ & 33.259/0.8418 & 29.645/0.7416 & 31.942/0.8069 & 32.722/0.8453 & \underline{34.511}/\underline{0.8694} & 33.698/0.8603 & 33.902/0.8614 & 30.063/0.7477 & \textbf{34.618}/\textbf{0.8790} \\
        $(1/9,180^\circ)$ & 36.059/0.8809 & 32.629/0.7888 & 35.703/0.8770 & 35.900/0.8834 & \underline{36.530}/\underline{0.8921} & 36.338/0.8886 & 36.352/0.8904 & 33.866/0.8429 & \textbf{36.632}/\textbf{0.8973} \\
        \midrule
        Mean & 37.093/0.8963 & 31.542/0.7911 & 36.217/0.8813 & 36.945/0.9001 & \textbf{38.390}/\underline{0.9171} & 37.863/0.9119 & 38.049/0.9137 & 32.658/0.8215 & \underline{38.362}/\textbf{0.9241} \\
        \bottomrule
    \end{tabular}}
\end{minipage}
\end{table*}

\begin{table*}[!t]
\centering
\begin{minipage}{\textwidth}
    \centering
    \captionof{table}{Results on TCIA Pancreas (3D). Each entry reports PSNR in dB followed by SSIM. The four setting groups are defined in Section~\ref{sec:exp_protocol}. Best and second-best values are shown in bold and underlined.}
    \label{tab:tcia_results}
    \scriptsize
    \setlength{\tabcolsep}{2.2pt}
    \renewcommand{\arraystretch}{0.98}
    \resizebox{\textwidth}{!}{%
    \begin{tabular}{@{}lccccccccc@{}}
        \toprule
        Setting & \shortstack{FBPConvNet\\TIP'17} & \shortstack{U-Net\\MICCAI'15} & \shortstack{FreeSeed\\MICCAI'23} & \shortstack{ProCT\\PR'26} & \shortstack{DOLCE\\ICCV'23} & \shortstack{NEED\\MIA'25} & \shortstack{TIFA\\TMI'24} & \shortstack{3DGR-CT\\MIA'25} & \shortstack{\method{}\\Ours} \\
        \midrule
        \multicolumn{10}{@{}l}{\textit{Limited-angle}} \\
        $90^\circ$ & 18.497/0.5746 & 19.973/\underline{0.6630} & 18.359/0.6067 & 19.593/0.5824 & 17.005/0.5251 & 19.997/0.6610 & 17.760/0.6055 & \underline{20.087}/0.6584 & \textbf{26.189}/\textbf{0.8500} \\
        $120^\circ$ & 21.835/0.6736 & \underline{23.632}/\underline{0.7469} & 21.841/0.6528 & 22.861/0.6590 & 20.207/0.6107 & 23.604/0.7374 & 21.404/0.6828 & 23.607/0.7364 & \textbf{28.507}/\textbf{0.8888} \\
        $150^\circ$ & 24.931/0.7294 & \underline{26.306}/\underline{0.7990} & 23.744/0.6407 & 25.605/0.6900 & 23.393/0.6625 & 26.199/0.7952 & 25.239/0.7379 & 26.284/0.7961 & \textbf{32.477}/\textbf{0.9357} \\
        $180^\circ$ & 28.339/0.8127 & \underline{30.565}/\underline{0.8528} & 25.763/0.6854 & 28.051/0.6736 & 26.612/0.6918 & 30.293/0.8101 & 29.862/0.7357 & 30.308/0.8294 & \textbf{45.999}/\textbf{0.9934} \\
        \midrule
        \multicolumn{10}{@{}l}{\textit{Sparse-view}} \\
        $1/3$ & 28.436/0.8140 & \underline{30.547}/\underline{0.8505} & 25.739/0.6846 & 28.042/0.6735 & 26.621/0.6920 & 30.285/0.8144 & 29.856/0.7358 & 30.355/0.8291 & \textbf{45.386}/\textbf{0.9912} \\
        $1/6$ & 28.277/0.8026 & \underline{30.604}/\underline{0.8500} & 25.567/0.6859 & 28.003/0.6737 & 26.558/0.6917 & 30.241/0.8167 & 29.817/0.7350 & 30.401/0.8319 & \textbf{43.281}/\textbf{0.9831} \\
        $1/9$ & 27.828/0.7635 & \underline{30.359}/\underline{0.8477} & 25.396/0.6846 & 27.856/0.6739 & 26.409/0.6881 & 30.090/0.8130 & 29.631/0.7320 & 30.203/0.8324 & \textbf{41.273}/\textbf{0.9759} \\
        $1/12$ & 27.251/0.7131 & \underline{29.993}/\underline{0.8402} & 25.307/0.6774 & 27.698/0.6716 & 26.263/0.6835 & 29.749/0.8115 & 29.349/0.7313 & 29.780/0.8232 & \textbf{39.735}/\textbf{0.9662} \\
        \midrule
        \multicolumn{10}{@{}l}{\textit{Sparse-view $+$ limited-angle}} \\
        $(1/3,90^\circ)$ & 18.487/0.5727 & 19.968/\underline{0.6656} & 18.327/0.6063 & 19.588/0.5826 & 17.015/0.5248 & 20.053/0.6609 & 17.775/0.6051 & \underline{20.060}/0.6598 & \textbf{26.050}/\textbf{0.8472} \\
        $(1/6,120^\circ)$ & 21.756/0.6588 & \underline{23.590}/\underline{0.7468} & 21.842/0.6585 & 22.795/0.6553 & 20.150/0.6072 & 23.554/0.7390 & 21.379/0.6799 & 23.573/0.7358 & \textbf{28.405}/\textbf{0.8824} \\
        $(1/9,150^\circ)$ & 24.557/0.6785 & \underline{26.290}/\underline{0.7974} & 23.806/0.6462 & 25.411/0.6763 & 23.270/0.6530 & 26.112/0.7903 & 25.132/0.7297 & 26.187/0.7889 & \textbf{31.455}/\textbf{0.9204} \\
        $(1/12,180^\circ)$ & 27.246/0.7134 & \underline{30.079}/\underline{0.8406} & 25.308/0.6774 & 27.698/0.6716 & 26.266/0.6835 & 29.716/0.8130 & 29.337/0.7311 & 29.793/0.8279 & \textbf{39.702}/\textbf{0.9660} \\
        \midrule
        \multicolumn{10}{@{}l}{\textit{Low-dose and joint degradation}} \\
        LD & 28.348/0.8025 & \underline{30.439}/\underline{0.8499} & 25.947/0.6915 & 28.224/0.6785 & 26.777/0.6953 & 30.082/0.8152 & 29.966/0.7333 & 30.288/0.8328 & \textbf{43.724}/\textbf{0.9830} \\
        \cmidrule(lr){1-10}
        $(1/3,120^\circ)$ & 18.727/0.2588 & \underline{22.729}/\underline{0.6895} & 20.016/0.3723 & 21.857/0.4861 & 19.123/0.3611 & 22.657/0.6863 & 20.264/0.4030 & 22.700/0.6821 & \textbf{27.801}/\textbf{0.8485} \\
        $(1/6,150^\circ)$ & 18.629/0.1949 & \underline{25.154}/0.7124 & 20.726/0.3202 & 23.842/0.4721 & 20.817/0.3352 & 24.990/\underline{0.7143} & 22.256/0.3954 & 25.012/0.7098 & \textbf{30.083}/\textbf{0.8771} \\
        $(1/9,180^\circ)$ & 18.445/0.2007 & \underline{27.415}/\underline{0.7226} & 20.909/0.3010 & 25.148/0.4522 & 21.535/0.3345 & 26.983/0.6757 & 23.421/0.4175 & 27.064/0.6878 & \textbf{34.705}/\textbf{0.9084} \\
        \midrule
        Mean & 23.849/0.6227 & \underline{26.728}/\underline{0.7797} & 23.037/0.5994 & 25.142/0.6233 & 23.001/0.5900 & 26.538/0.7596 & 25.153/0.6494 & 26.606/0.7664 & \textbf{35.298}/\textbf{0.9261} \\
        \bottomrule
    \end{tabular}}
\end{minipage}
\end{table*}

\subsection{3D reconstruction on LDCT-PD}

Table~\ref{tab:diffct_results} shows that \method{} achieves the highest PSNR and SSIM in every LDCT-PD setting. Its macro-average reaches 32.655 dB and 0.8661 SSIM, improving over the strongest metric-specific comparators by 5.808 dB and 0.1131 SSIM. The gains span all corruption groups. Under limited-angle acquisition, the PSNR improvement over the strongest baseline ranges from 4.709 dB at $90^{\circ}$ to 11.067 dB at $180^{\circ}$. Under sparse-view acquisition, \method{} retains 35.044 dB and 0.8834 SSIM at the most severe $1/12$ view ratio.

The paired sparse-view and limited-angle results further show that the model remains stable when angular density and angular support are reduced together. The most demanding group additionally introduces low-dose noise. Averaged over its three joint rows, \method{} obtains 29.067 dB and 0.8014 SSIM. It exceeds the strongest comparator averages by 3.301 dB and 0.0707 SSIM. This persistent margin under coupled corruption is notable because the remaining measurements contain both directional incompleteness and signal-dependent noise.

\begin{figure*}[!t]
    \centering
    \includegraphics[pagebox=cropbox,width=\textwidth]{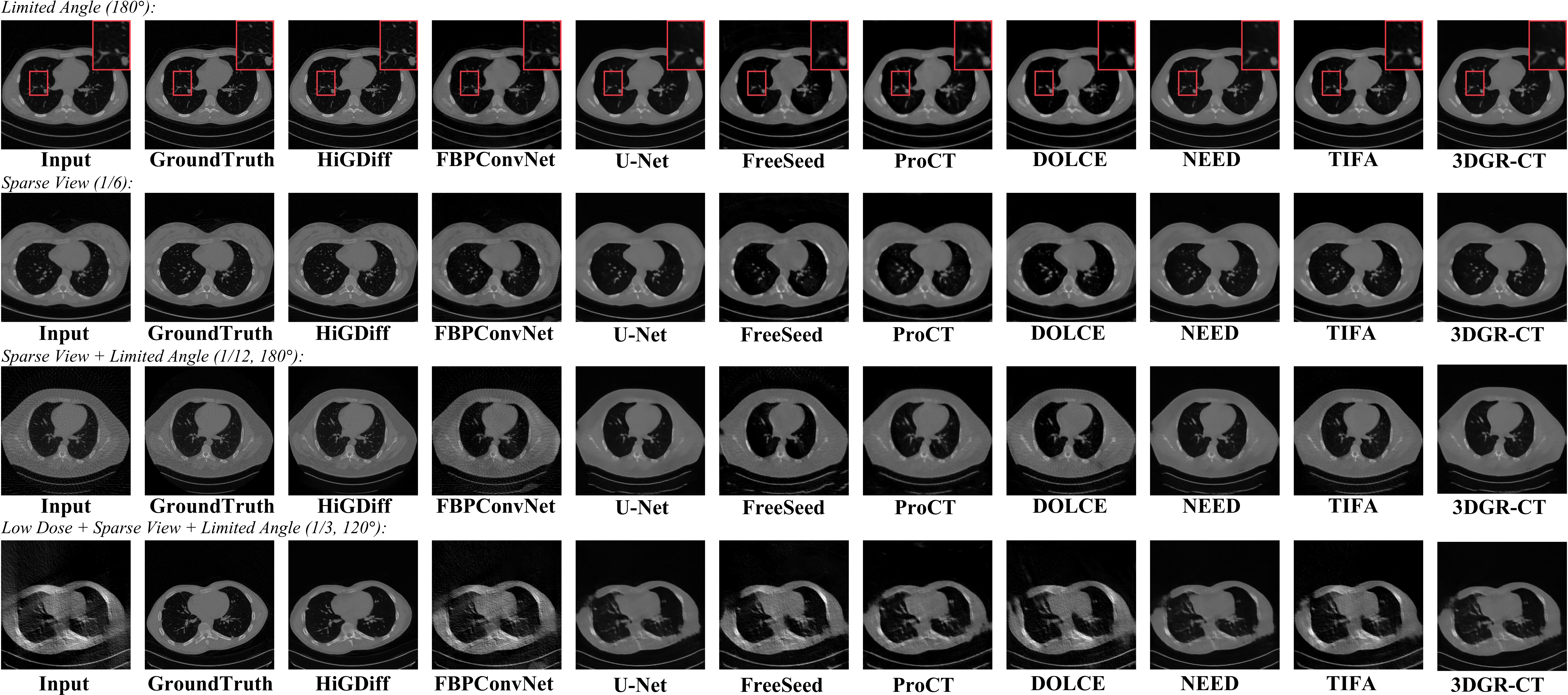}
    \caption{Qualitative comparison on LDCT-PD under representative limited-angle, sparse-view, paired sparse-view and limited-angle, and joint low-dose settings. Red boxes identify regions enlarged for boundary and artifact comparison. All methods are displayed using the same attenuation window within each row. The joint row shows mild low-contrast boundary smoothing.}
    \label{fig:qual_diffct}
\end{figure*}

Figure~\ref{fig:qual_diffct} provides the corresponding visual comparison at representative limited-angle, sparse-view, paired, and joint settings. In the limited-angle example, competing reconstructions retain slope artifacts or attenuate fine boundaries. \method{} recovers a more continuous thoracic structure. The same pattern persists as sparse sampling is added, with distributed streaks suppressed and local contrast retained. Under joint degradation, \method{} remains visually closest to the reference and reduces the structured artifacts and tissue smoothing visible in the other reconstructions.

\subsection{2D reconstruction on LoDoPaB-CT}

Table~\ref{tab:lodopab_results} presents a more competitive 2D benchmark. DOLCE obtains the highest macro-average PSNR at 38.390 dB, 0.028 dB above \method{}. \method{} gives the highest mean SSIM at 0.9241, exceeding DOLCE by 0.0070. At the setting level, \method{} ranks first in 10 of 16 PSNR comparisons and 14 of 16 SSIM comparisons. NEED and DOLCE remain stronger in several mild limited-angle and sparse-view settings, so PSNR superiority depends on the acquisition setting.

The relative ordering changes under compound degradation. \method{} ranks first in both metrics for all three sparse-view, limited-angle, and low-dose rows, averaging 34.715 dB and 0.8799 SSIM. The gains over the strongest comparator averages are 0.117 dB and 0.0088 SSIM. On LoDoPaB-CT, these results show improved structural fidelity and robustness to coupled corruption, alongside a macro-average PSNR effectively tied with the strongest diffusion baseline.

\subsection{Cross-dataset generalization on TCIA Pancreas}

Table~\ref{tab:tcia_results} evaluates cross-dataset generalization from LDCT-PD to TCIA Pancreas. \method{} achieves 35.298 dB and 0.9261 SSIM on average. It ranks first for both metrics in all 16 settings, outperforming the strongest comparator averages by 8.571 dB and 0.1464 SSIM. The gains across both limited-angle and sparse-view groups indicate that the learned representation generalizes across the anatomical shift from thorax to abdomen.

Under the three joint protocols, \method{} averages 30.863 dB and 0.8780 SSIM, exceeding the strongest comparator averages by 5.764 dB and 0.1698 SSIM.

\begin{figure*}[!t]
    \centering
    \includegraphics[pagebox=cropbox,width=\textwidth]{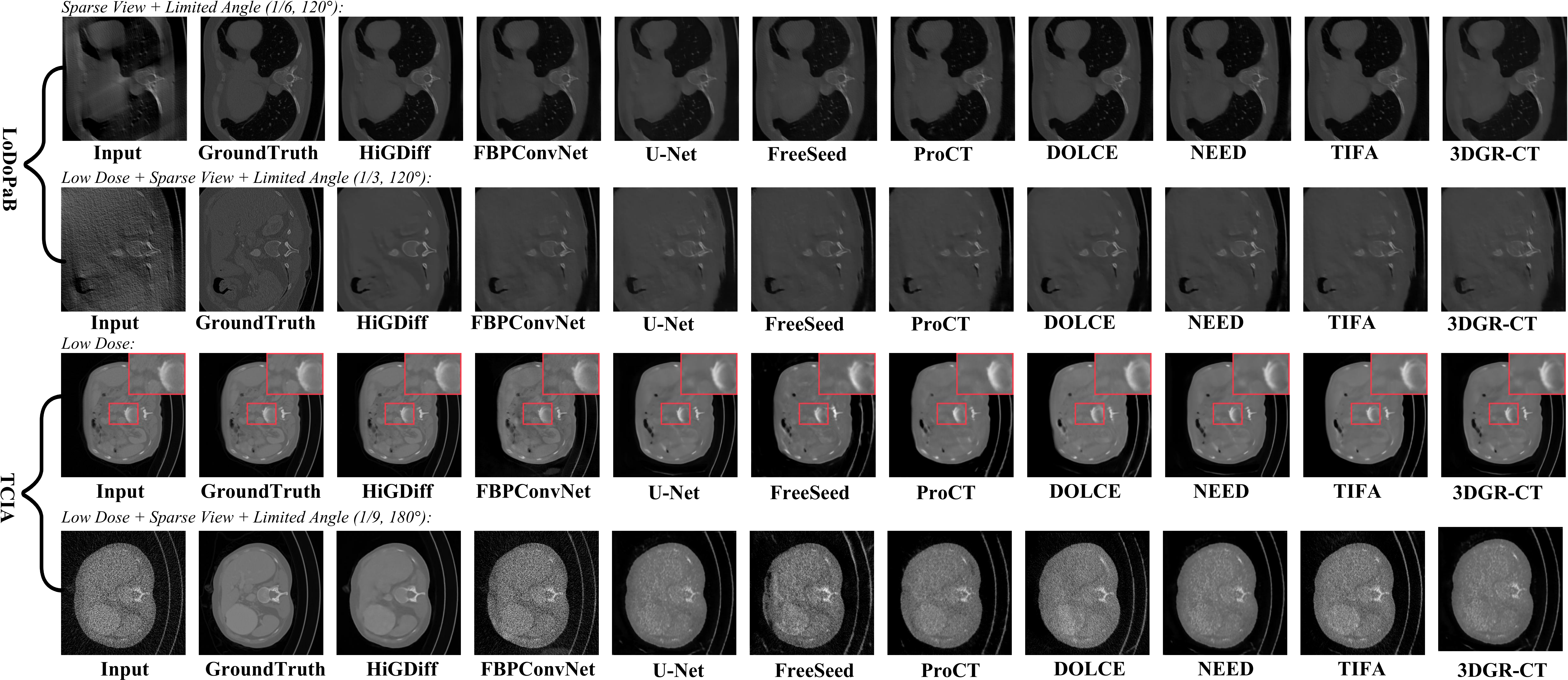}
    \caption{Qualitative comparison on LoDoPaB-CT in the top two rows and TCIA Pancreas in the bottom two rows. The examples cover compound sparse-view and limited-angle degradation and low-dose combinations. Residual directional texture and low-contrast smoothing remain visible under the most severe combinations.}
    \label{fig:qual_cross_dataset}
\end{figure*}

The cross-dataset examples in Figure~\ref{fig:qual_cross_dataset} complement these measurements. On LoDoPaB-CT, where the numerical comparison is close, \method{} preserves structural continuity under compound degradation and limits amplification of low-dose noise. On TCIA Pancreas, it reduces residual noise and directional artifacts around abdominal structures. These images illustrate the observed generalization trend.

\subsection{Ablation studies}
\label{sec:exp_ablation}

We conduct the ablations on LDCT-PD and average PSNR and SSIM over the 16 fixed acquisition settings. The study follows the three methodological components in Sections~\ref{sec:init}--\ref{sec:refinement}. They are foreground-aware initialization, structure-to-detail Gaussian diffusion, and gradient-isolated attenuation refinement. We additionally examine the grouped objectives in Section~\ref{sec:training}, visualize successive reconstruction stages, and report reverse-step sensitivity.

\subsubsection{Core components}

Table~\ref{tab:component_ablation} examines the initialization and refinement modules that support the hierarchical design. Removing the physics residual from $F_{\mathrm{phy}}$, the anchor set $\mathcal K$, or the foreground field $F_{\mathrm{fg}}$ reduces performance, with the anchors producing the largest PSNR loss among the three initialization cues. Removing $H$ evaluates the dual-bank Gaussian rendering directly. Its 25.471 dB and 0.5854 SSIM show that the Gaussian hierarchy alone recovers a meaningful 3D attenuation field, while a finite set of smooth primitives leaves voxel-scale error at sharp tissue interfaces and under structured streak and noise corruption. $H$ operates on the dense 3D attenuation grid and combines $\hat\mu_G$, the residual backprojection $B_G$, and $F_{\mathrm{fg}}$ to correct these errors with a bounded multiscale residual. Relative to Gaussian rendering alone, $H$ contributes a further 7.184 dB and 0.2807 SSIM, quantifying the complementary contribution of dense voxel refinement.

The hierarchy remains necessary when $H$ is present. All variants in Table~\ref{tab:structure_ablation} retain the same refinement network and training schedule. Removing bank decomposition reduces PSNR from 32.655 to 30.813 dB, removing sequential conditioning reduces it to 30.146 dB, and replacing Gaussian diffusion with direct correction prediction reduces it to 24.832 dB. Since $H$ is unchanged in these comparisons, the remaining differences arise from the Gaussian representation and its structure-to-detail inference. The two stages have complementary roles: the Gaussian banks organize the recoverable 3D attenuation field, and $H$ completes the residual information on the voxel grid. Allowing the refinement loss to cross the stop-gradient boundary also lowers both metrics, supporting the staged optimization of these roles.

\par\medskip
\begin{center}
\begin{minipage}{0.96\columnwidth}
    \centering
    \captionof{table}{Core component ablation on LDCT-PD.}
    \label{tab:component_ablation}
    \small
    \setlength{\tabcolsep}{7pt}
    \begin{tabular}{@{}lcc@{}}
        \toprule
        Variant & PSNR (dB) $\uparrow$ & SSIM $\uparrow$ \\
        \midrule
        w/o physics residual & 24.584 & 0.7449 \\
        w/o anchors $\mathcal K$ & 29.317 & 0.7680 \\
        w/o $F_{\mathrm{fg}}$ & 30.372 & 0.8336 \\
        w/o refinement $H$ & 25.471 & 0.5854 \\
        w/o $\sg(\cdot)$ & 30.747 & 0.8046 \\
        \midrule
        \textbf{Full \method{}} & \textbf{32.655} & \textbf{0.8661} \\
        \bottomrule
    \end{tabular}
\end{minipage}
\end{center}
\par\medskip

\subsubsection{Hierarchical structure}

Table~\ref{tab:structure_ablation} isolates the organization of the Gaussian banks and their conditioning paths while retaining $H$. Removing the structure and detail bank decomposition or their sequential dependency lowers both metrics. The complete hierarchy outperforms both single-condition variants, showing that $R_s$ and $B_s$ provide complementary information. Replacing DDIM diffusion with direct correction prediction also degrades performance, supporting iterative Gaussian denoising for the slot-aligned residuals.

\begin{center}
\begin{minipage}{0.96\columnwidth}
    \centering
    \captionof{table}{Ablation of the structure-to-detail design on LDCT-PD.}
    \label{tab:structure_ablation}
    \small
    \setlength{\tabcolsep}{7pt}
    \begin{tabular}{@{}lcc@{}}
        \toprule
        Variant & PSNR (dB) $\uparrow$ & SSIM $\uparrow$ \\
        \midrule
        w/o bank decomposition & 30.813 & 0.8443 \\
        w/o sequential conditioning & 30.146 & 0.8440 \\
        w/o structure tokens $R_s$ & 30.859 & 0.8450 \\
        w/o residual evidence $B_s$ & 30.792 & 0.8436 \\
        w/o DDIM diffusion & 24.832 & 0.7395 \\
        \midrule
        \textbf{Full hierarchy} & \textbf{32.655} & \textbf{0.8661} \\
        \bottomrule
    \end{tabular}
\end{minipage}
\end{center}

\subsubsection{Loss functions}

Table~\ref{tab:loss_ablation} shows that the structure and detail separation objective has the clearest effect on SSIM, and the residual penalty is more influential for PSNR. Removing the auxiliary or projection objective produces smaller differences among the reduced variants. These terms mainly stabilize anchor and foreground learning and observed-ray fidelity.

\begin{center}
\begin{minipage}{0.96\columnwidth}
    \centering
    \captionof{table}{Ablation of the grouped objectives on LDCT-PD.}
    \label{tab:loss_ablation}
    \small
    \setlength{\tabcolsep}{8pt}
    \begin{tabular}{@{}lcc@{}}
        \toprule
        Objective & PSNR (dB) $\uparrow$ & SSIM $\uparrow$ \\
        \midrule
        w/o $\mathcal L_{\mathrm{aux}}$ & 30.846 & 0.8436 \\
        w/o $\mathcal L_{\mathrm{sep}}$ & 30.799 & 0.8280 \\
        w/o $\mathcal L_{\mathrm{proj}}$ & 30.816 & 0.8445 \\
        w/o $\mathcal L_{\mathrm{res}}$ & 30.689 & 0.8431 \\
        \midrule
        \textbf{Full objective} & \textbf{32.655} & \textbf{0.8661} \\
        \bottomrule
    \end{tabular}
\end{minipage}
\end{center}

\subsubsection{Output stages and reverse steps}

\begin{figure}[!t]
    \centering
    \includegraphics[pagebox=cropbox,width=\linewidth]{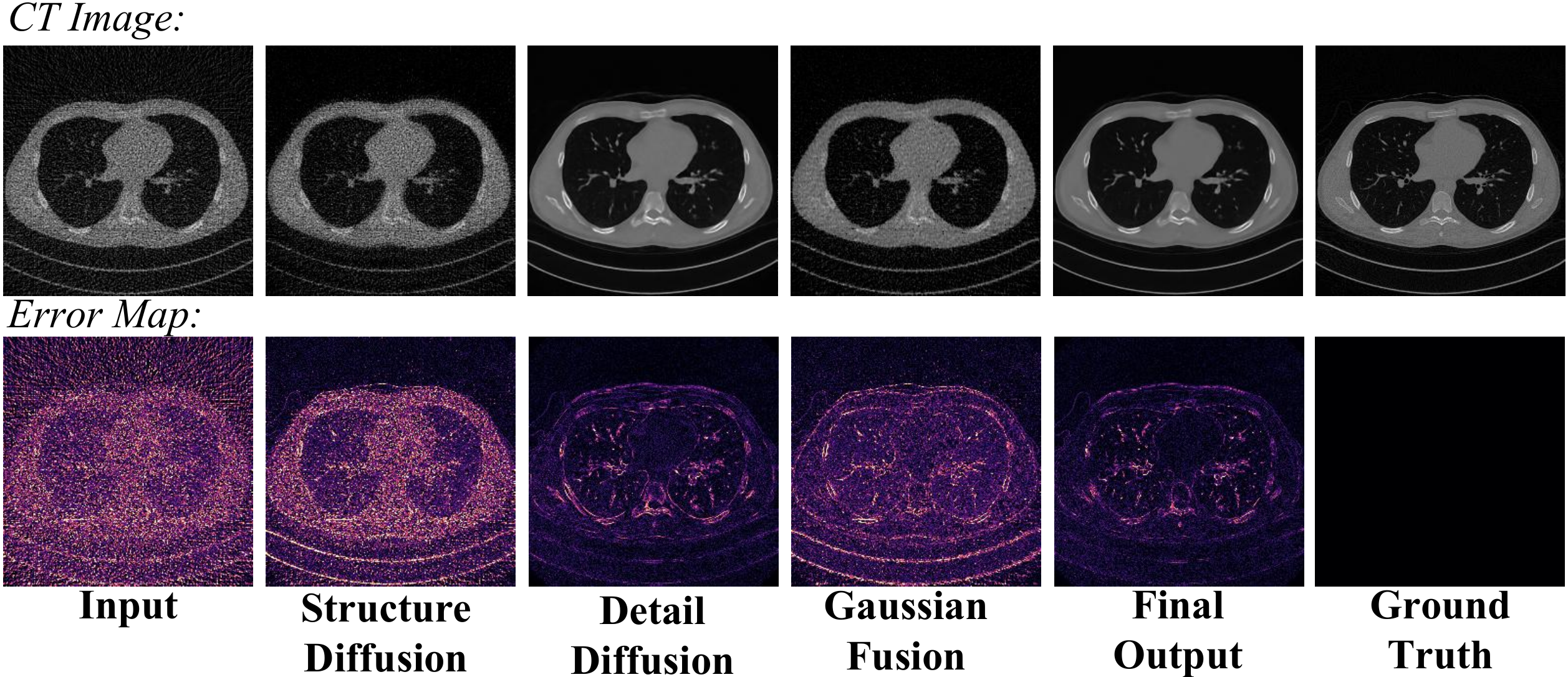}
    \caption{Qualitative evolution of the HiGDiff reconstruction on LDCT-PD. The upper row shows the output of each stage, and the lower row shows its absolute error map relative to the ground truth.}
    \label{fig:stage_outputs}
\end{figure}

Figure~\ref{fig:stage_outputs} visualizes the successive outputs of the reconstruction path. Structure diffusion establishes the coarse attenuation organization, after which detail diffusion and Gaussian fusion sharpen local boundaries and reduce structured error. The final bounded refinement removes residual voxel-domain discrepancies and retains the Gaussian reconstruction as its base. Table~\ref{tab:step_ablation} further shows a monotonic gain as the deterministic schedules increase from one to three reverse steps.

\begin{center}
\begin{minipage}{0.96\columnwidth}
    \centering
    \captionof{table}{Reverse-step sensitivity on LDCT-PD, reported as averages over the complete 16-setting protocol.}
    \label{tab:step_ablation}
    \small
    \setlength{\tabcolsep}{9pt}
    \begin{tabular}{@{}lcc@{}}
        \toprule
        $(T_s,T_d)$ & PSNR (dB) $\uparrow$ & SSIM $\uparrow$ \\
        \midrule
        $(1,1)$ & 30.147 & 0.8491 \\
        $(2,2)$ & 31.260 & 0.8639 \\
        $(3,3)$ (full) & \textbf{32.655} & \textbf{0.8661} \\
        \bottomrule
    \end{tabular}
\end{minipage}
\end{center}

\section{Discussion}
\label{sec:discussion}

\subsection{Clinical Relevance of Joint Degradation}

Reducing view count, angular coverage, and photon flux simultaneously represents a clinically relevant stress regime when acquisition time, scanning trajectories, and radiation exposure are jointly constrained. The resulting errors are coupled. Limited angles remove direction-dependent information and produce slope artifacts, sparse sampling distributes streaks, and low photon counts corrupt the remaining evidence. The joint quantitative results and the visual comparisons in Figures~\ref{fig:qual_diffct} and~\ref{fig:qual_cross_dataset} indicate that the structure-to-detail representation can preserve both long-range attenuation organization and local tissue transitions under this combined loss of information. Improved PSNR, SSIM, and visual fidelity indicate potential value for dose- and time-constrained CT.

\subsection{Low-Data Cross-Dataset Generalization}

The LDCT-PD-to-TCIA experiment evaluates transfer from thoracic LDCT-PD to abdominal TCIA Pancreas. The results show that the Gaussian hierarchy transfers effectively across anatomy and acquisition statistics, while target-domain adaptation calibrates dataset-specific intensity and noise.

The TCIA ranking also provides insight into low-data generalization. U-Net \citep{ronneberger2015u} is the strongest comparator in macro-average PSNR and SSIM and ranks second in most individual settings. Its relatively small parameter count facilitates adaptation with limited target-domain data, whereas the larger parameter spaces of the other reconstruction models require more data for reliable adaptation. The functionally distinct Gaussian banks provide a complementary adaptation mechanism. The structure bank preserves transferable global attenuation organization, while the detail bank reallocates representation capacity to target-domain tissue transitions. This structured representation supports generalization to a new anatomy with limited target-domain data.

\subsection{Limitations and Future Work}

\begin{figure}[!t]
    \centering
    \includegraphics[pagebox=cropbox,width=\linewidth]{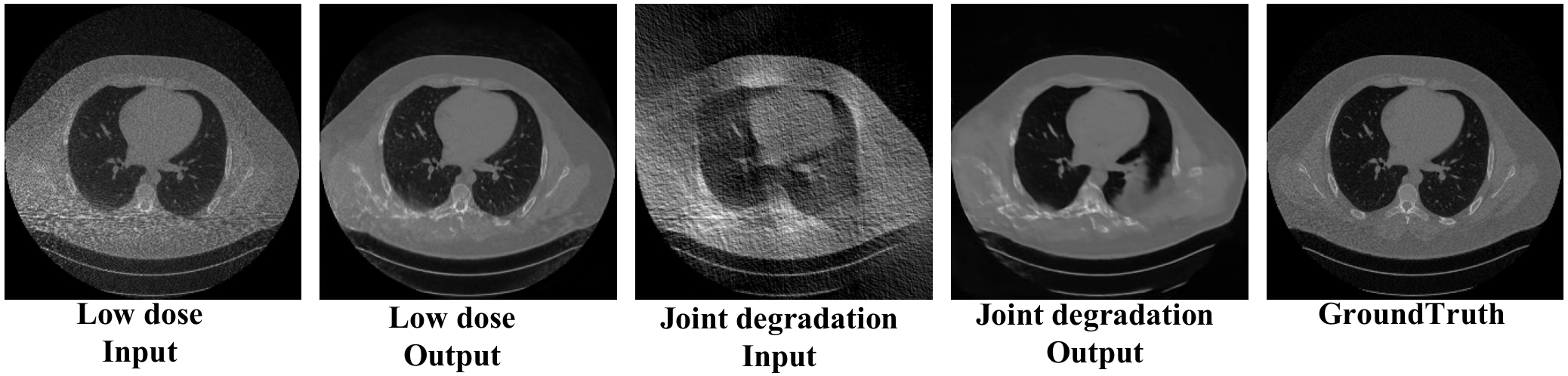}
    \caption{Representative failure case. The columns show the low-dose input and output, joint-degradation input and output, and ground truth. Severe joint degradation leaves residual smoothing and incomplete low-contrast structure.}
    \label{fig:failure_case}
\end{figure}

The most severe joint settings expose the current operating boundary. On LDCT-PD, performance decreases to 27.172 dB and 0.7781 SSIM at the joint $(1/3,120^{\circ})$ low-dose setting, compared with the macro-average of 32.655 dB and 0.8661 SSIM. The corresponding difficult examples in Figures~\ref{fig:qual_diffct} and~\ref{fig:qual_cross_dataset} retain mild boundary smoothing and residual directional texture in low-contrast regions. Figure~\ref{fig:failure_case} further illustrates this limitation on a representative thoracic slice. The low-dose reconstruction recovers most gross anatomy. Joint degradation introduces severe directional streaks and photon noise, and the reconstruction suppresses much of this corruption. Portions of the low-contrast thoracic structure remain incompletely recovered. Information lost to the angular null space and photon noise remains fundamentally ambiguous, even under a strong learned prior.

The evaluated degradations rely substantially on simulated or protocol-derived projection corruption. Consistency on observed rays leaves the limited-angle null space unresolved, so subtle low-contrast findings may remain uncertain. Raw photon-count data, prospective acquisitions, multi-centre validation, attenuation-bias analysis, task-based assessment, and blinded reader evaluation are therefore required before clinical conclusions.

The fixed Gaussian budget and structure and detail bandwidth may also underrepresent very small or low-contrast structures. Adaptive primitive allocation, learned multiscale decomposition, and uncertainty-aware capacity control are natural extensions. Fixed DDIM sampling \citep{song2021ddim} introduces iterative computation. Future work will investigate self-supervised fine-tuning, domain-generalized training, faster samplers, and explicit analysis of reconstruction quality and runtime.

\section{Conclusion}
\label{sec:conclusion}

This work presented \method{} for 3D CT reconstruction under severely constrained acquisition across isolated, paired, and joint degradation settings. \method{} decomposes the resulting ambiguity spatially through foreground-aware Gaussian allocation and hierarchically through structure-conditioned detail diffusion. The two functionally distinct Gaussian banks recover global attenuation organization and residual tissue transitions. Gradient-isolated refinement corrects the rendered field and preserves the learned primitive parameters. Experiments across LDCT-PD, LoDoPaB-CT, and TCIA Pancreas demonstrate consistent gains under joint degradation, and the TCIA study shows cross-dataset generalization. Together, these findings support hierarchical Gaussian diffusion as an explicit and adaptable representation for severely constrained CT reconstruction.

\creditauthor{Yuezhe Yang}{Conceptualization, Methodology, Software, Validation, Formal analysis, Investigation, Data curation, Visualization, Writing -- original draft, Writing -- review and editing}
\creditauthor{Li Cheng}{Supervision, Project administration, Writing -- review and editing}
\printcredits

\section*{Acknowledgements}
Yuezhe Yang acknowledges support from the China Scholarship Council (CSC).
We acknowledge the creators of the LDCT-and-Projection-data collection and the support of NIBIB grants EB017095 and EB017185 (Cynthia McCollough, PI) for its creation.

\bibliographystyle{cas-model2-names}
\setlength{\bibsep}{0pt plus 0.2ex}
\bibliography{cas-refs}

@inproceedings{mayo1991high,
  title={The high-resolution computed tomography technique},
  author={Mayo, John R},
  booktitle={Seminars in roentgenology},
  volume={26},
  number={2},
  pages={104--109},
  year={1991},
  organization={Elsevier}
}

@article{adams2009quantitative,
  title={Quantitative computed tomography},
  author={Adams, Judith E},
  journal={European journal of radiology},
  volume={71},
  number={3},
  pages={415--424},
  year={2009},
  publisher={Elsevier}
}

@article{liang2017guest,
  title={Guest editorial low-dose CT: what has been done, and what challenges remain?},
  author={Liang, Jerome Zhengrong and La Riviere, Patrick J and El Fakhri, Georges and Glick, Stephen J and Siewerdsen, Jeff},
  journal={IEEE Transactions on Medical Imaging},
  volume={36},
  number={12},
  pages={2409--2416},
  year={2017},
  publisher={IEEE}
}

@article{barrett2004artifacts,
  title={Artifacts in CT: recognition and avoidance},
  author={Barrett, Julia F and Keat, Nicholas},
  journal={Radiographics},
  volume={24},
  number={6},
  pages={1679--1691},
  year={2004},
  publisher={Radiological Society of North America}
}

@article{deak2013filtered,
  title={Filtered back projection, adaptive statistical iterative reconstruction, and a model-based iterative reconstruction in abdominal CT: an experimental clinical study},
  author={De{\'a}k, Zsuzsanna and Grimm, Jochen M and Treitl, Marcus and Geyer, Lucas L and Linsenmaier, Ulrich and K{\"o}rner, Markus and Reiser, Maximilian F and Wirth, Stefan},
  journal={Radiology},
  volume={266},
  number={1},
  pages={197--206},
  year={2013},
  publisher={Radiological Society of North America, Inc.}
}

@article{koetzier2023deep,
  title={Deep learning image reconstruction for CT: technical principles and clinical prospects},
  author={Koetzier, Lennart R and Mastrodicasa, Domenico and Szczykutowicz, Timothy P and van der Werf, Niels R and Wang, Adam S and Sandfort, Veit and van der Molen, Aart J and Fleischmann, Dominik and Willemink, Martin J},
  journal={Radiology},
  volume={306},
  number={3},
  pages={e221257},
  year={2023},
  publisher={Radiological Society of North America}
}

@inproceedings{ronneberger2015u,
  title={U-net: Convolutional networks for biomedical image segmentation},
  author={Ronneberger, Olaf and Fischer, Philipp and Brox, Thomas},
  booktitle={International Conference on Medical image computing and computer-assisted intervention},
  pages={234--241},
  year={2015},
  organization={Springer}
}

@inproceedings{ma2023freeseed,
  title={FreeSeed: Frequency-band-aware and self-guided network for sparse-view CT reconstruction},
  author={Ma, Chenglong and Li, Zilong and Zhang, Junping and Zhang, Yi and Shan, Hongming},
  booktitle={International conference on medical image computing and computer-assisted intervention},
  pages={250--259},
  year={2023},
  organization={Springer}
}

@article{jin2017deep,
  title={Deep convolutional neural network for inverse problems in imaging},
  author={Jin, Kyong Hwan and McCann, Michael T and Froustey, Emmanuel and Unser, Michael},
  journal={IEEE transactions on image processing},
  volume={26},
  number={9},
  pages={4509--4522},
  year={2017},
  publisher={IEEE}
}

@article{gao2025noise,
  title={Noise-Inspired Diffusion Model for Generalizable Low-Dose CT Reconstruction},
  author={Gao, Qi and Chen, Zhihao and Zeng, Dong and Zhang, Junping and Ma, Jianhua and Shan, Hongming},
  journal={Medical Image Analysis},
  volume={105},
  pages={103710},
  year={2025},
  doi={10.1016/j.media.2025.103710}
}

@inproceedings{liu2023dolce,
  title={Dolce: A model-based probabilistic diffusion framework for limited-angle ct reconstruction},
  author={Liu, Jiaming and Anirudh, Rushil and Thiagarajan, Jayaraman J and He, Stewart and Mohan, K Aditya and Kamilov, Ulugbek S and Kim, Hyojin},
  booktitle={Proceedings of the IEEE/CVF international conference on computer vision},
  pages={10498--10508},
  year={2023}
}

@article{yang2026representation,
  title={Representation Paradigms in AI-based 3D Radiological Image Reconstruction: A Systematic Review},
  author={Yang, Yuezhe and Bi, Lei and Yang, Boyu and Wang, Yaqian and He, Yang and Peng, Yige and Jin, Zhe and Dong, Xingbo and Kim, Jinman},
  journal={arXiv preprint arXiv:2504.11349},
  year={2026},
  doi={10.48550/arXiv.2504.11349}
}

@article{chen2017redcnn,
  title={Low-Dose {CT} With a Residual Encoder--Decoder Convolutional Neural Network},
  author={Chen, Hu and Zhang, Yi and Kalra, Mannudeep K. and Lin, Feng and Chen, Yang and Liao, Peixi and Zhou, Jiliu and Wang, Ge},
  journal={IEEE Transactions on Medical Imaging},
  volume={36},
  number={12},
  pages={2524--2535},
  year={2017},
  doi={10.1109/TMI.2017.2715284}
}

@article{chen2018learn,
  title={{LEARN}: Learned Experts' Assessment-Based Reconstruction Network for Sparse-Data {CT}},
  author={Chen, Hu and Zhang, Yi and Chen, Yunjin and Zhang, Junfeng and Zhang, Weihua and Sun, Huaiqiang and Lv, Yang and Liao, Peixi and Zhou, Jiliu and Wang, Ge},
  journal={IEEE Transactions on Medical Imaging},
  volume={37},
  number={6},
  pages={1333--1347},
  year={2018},
  doi={10.1109/TMI.2018.2805692}
}

@article{ma2024proct,
  title={Universal Pre-training for Generalizable Incomplete-View {CT} Reconstruction},
  author={Ma, Chenglong and Li, Zilong and He, Junjun and Zhang, Junping and Zhang, Yi and Shan, Hongming},
  journal={Pattern Recognition},
  volume={178},
  pages={113513},
  year={2026},
  doi={10.1016/j.patcog.2026.113513}
}

@article{wang2024tifa,
  title={Time-Reversion Fast-Sampling Score-Based Model for Limited-Angle {CT} Reconstruction},
  author={Wang, Yanyang and Li, Zirong and Wu, Weiwen},
  journal={IEEE Transactions on Medical Imaging},
  volume={43},
  number={10},
  pages={3449--3460},
  year={2024},
  doi={10.1109/TMI.2024.3418838}
}

@article{leuschner2021lodopab,
  title={{LoDoPaB-CT}, a Benchmark Dataset for Low-Dose Computed Tomography Reconstruction},
  author={Leuschner, Johannes and Schmidt, Maximilian and Otero Baguer, Daniel and Maass, Peter},
  journal={Scientific Data},
  volume={8},
  pages={109},
  year={2021},
  doi={10.1038/s41597-021-00893-z}
}

@misc{mccollough2020ldct,
  title={Low Dose {CT} Image and Projection Data ({LDCT-and-Projection-data}) (Version 7)},
  author={McCollough, Cynthia and Chen, Bin and Holmes, David R. and Duan, Xiaohui and Yu, Zhiqiang and Yu, Lifeng and Leng, Shuai and Fletcher, Joel G.},
  howpublished={The Cancer Imaging Archive},
  year={2020},
  doi={10.7937/9NPB-2637}
}

@misc{roth2016pancreasct,
  title={Data From {Pancreas-CT} (Version 2)},
  author={Roth, Holger and Farag, Amal and Turkbey, Evrim B. and Lu, Le and Liu, Jiamin and Summers, Ronald M.},
  howpublished={The Cancer Imaging Archive},
  year={2016},
  doi={10.7937/K9/TCIA.2016.tNB1kqBU}
}

@inproceedings{roth2015deeporgan,
  title={{DeepOrgan}: Multi-level deep convolutional networks for automated pancreas segmentation},
  author={Roth, Holger R. and Lu, Le and Farag, Amal and Shin, Hoo-Chang and Liu, Jiamin and Turkbey, Evrim B. and Summers, Ronald M.},
  booktitle={Medical Image Computing and Computer-Assisted Intervention -- MICCAI 2015},
  series={Lecture Notes in Computer Science},
  volume={9349},
  pages={556--564},
  year={2015},
  publisher={Springer},
  doi={10.1007/978-3-319-24553-9_68}
}

@article{clark2013tcia,
  title={The Cancer Imaging Archive ({TCIA}): Maintaining and operating a public information repository},
  author={Clark, Kenneth and Vendt, Bruce and Smith, Kirk and Freymann, John and Kirby, Justin and Koppel, Paul and Moore, Stephen and Phillips, Stanley and Maffitt, David and Pringle, Michael and Tarbox, Lawrence and Prior, Fred},
  journal={Journal of Digital Imaging},
  volume={26},
  number={6},
  pages={1045--1057},
  year={2013},
  doi={10.1007/s10278-013-9622-7}
}

@misc{leuschner2019lodopabdataset,
  title={{LoDoPaB-CT} Dataset (Version 1.0.0)},
  author={Leuschner, Johannes and Schmidt, Maximilian and Otero Baguer, Daniel},
  howpublished={Zenodo},
  year={2019},
  doi={10.5281/zenodo.3384092}
}

@article{kerbl20233d,
  title={3D Gaussian splatting for real-time radiance field rendering.},
  author={Kerbl, Bernhard and Kopanas, Georgios and Leimk{\"u}hler, Thomas and Drettakis, George},
  journal={ACM Trans. Graph.},
  volume={42},
  number={4},
  pages={139--1},
  year={2023}
}

@article{li20253dgr,
  title={3DGR-CT: Sparse-view CT reconstruction with a 3D Gaussian representation},
  author={Li, Yingtai and Fu, Xueming and Li, Han and Zhao, Shang and Jin, Ruiyang and Zhou, S Kevin},
  journal={Medical Image Analysis},
  volume={103},
  pages={103585},
  year={2025},
  doi={10.1016/j.media.2025.103585},
  publisher={Elsevier}
}

@inproceedings{you2022ukpgan,
  title={{UKPGAN}: A General Self-Supervised Keypoint Detector},
  author={You, Yang and Liu, Wenhai and Ze, Yanjie and Li, Yong-Lu and Wang, Weiming and Lu, Cewu},
  booktitle={Proceedings of the IEEE/CVF Conference on Computer Vision and Pattern Recognition},
  pages={17042--17051},
  year={2022}
}

@inproceedings{zhou2024diffgs,
  title={{DiffGS}: Functional Gaussian Splatting Diffusion},
  author={Zhou, Junsheng and Zhang, Weiqi and Liu, Yu-Shen},
  booktitle={Advances in Neural Information Processing Systems},
  volume={37},
  year={2024},
  doi={10.52202/079017-1185}
}

@inproceedings{shi2026registers,
  title={Vision Transformers Need More Than Registers},
  author={Shi, Cheng and Yu, Yizhou and Yang, Sibei},
  booktitle={Proceedings of the IEEE/CVF Conference on Computer Vision and Pattern Recognition},
  pages={26328--26337},
  year={2026}
}

@inproceedings{singh2024dcdm,
  title={{DCDM}: Diffusion-Conditioned-Diffusion Model for Scene Text Image Super-Resolution},
  author={Singh, Shrey and Keserwani, Prateek and Iwamura, Masakazu and Roy, Partha Pratim},
  booktitle={European Conference on Computer Vision},
  pages={1474--1484},
  year={2024}
}

@inproceedings{lin2024difgaussian,
  title={Learning 3D Gaussians for Extremely Sparse-View Cone-Beam CT Reconstruction},
  author={Lin, Yiqun and Wang, Hualiang and Chen, Jixiang and Li, Xiaomeng},
  booktitle={Medical Image Computing and Computer Assisted Intervention -- MICCAI 2024},
  series={Lecture Notes in Computer Science},
  volume={15007},
  pages={425--435},
  year={2024},
  doi={10.1007/978-3-031-72104-5_41}
}

@inproceedings{chung2023diffusionmbir,
  title={Solving 3D Inverse Problems Using Pre-Trained 2D Diffusion Models},
  author={Chung, Hyungjin and Ryu, Dohoon and McCann, Michael T. and Klasky, Marc L. and Ye, Jong Chul},
  booktitle={Proceedings of the IEEE/CVF Conference on Computer Vision and Pattern Recognition},
  pages={22542--22551},
  year={2023}
}

@article{baguer2022untrained,
  title={Sparse-view and limited-angle CT reconstruction with untrained networks and deep image prior},
  author={Shu, Ziyu and Entezari, Alireza},
  journal={Computer Methods and Programs in Biomedicine},
  volume={226},
  pages={107167},
  year={2022},
  doi={10.1016/j.cmpb.2022.107167}
}

@article{li2025ddoct,
  title={{DDoCT}: Morphology preserved dual-domain joint optimization for fast sparse-view low-dose CT imaging},
  author={Li, Linxuan and Zhang, Zhijie and Li, Yongqing and Wang, Yanxin and Zhao, Wei},
  journal={Medical Image Analysis},
  volume={101},
  pages={103420},
  year={2025},
  doi={10.1016/j.media.2024.103420}
}

@article{yang2025ctsdm,
  title={{CT-SDM}: A Sampling Diffusion Model for Sparse-View CT Reconstruction Across Various Sampling Rates},
  author={Yang, Liutao and Huang, Jiahao and Yang, Guang and Zhang, Daoqiang},
  journal={IEEE Transactions on Medical Imaging},
  volume={44},
  pages={2581--2593},
  year={2025},
  doi={10.1109/TMI.2025.3541491}
}

@article{chen2025cddm,
  title={Mitigating Data Consistency Induced Discrepancy in Cascaded Diffusion Models for Sparse-View CT Reconstruction},
  author={Chen, Hanyu and Hao, Zhixiu and Guo, Lin and Xiao, Liying},
  journal={IEEE Transactions on Medical Imaging},
  volume={44},
  number={7},
  year={2025},
  doi={10.1109/TMI.2025.3557243}
}

@inproceedings{chen2025cvgdiff,
  title={Cross-view Generalized Diffusion Model for Sparse-view CT Reconstruction},
  author={Chen, Jixiang and Lin, Yiqun and Qin, Yi and Wang, Hualiang and Li, Xiaomeng},
  booktitle={Medical Image Computing and Computer Assisted Intervention -- MICCAI 2025},
  series={Lecture Notes in Computer Science},
  volume={15975},
  year={2025},
  publisher={Springer Nature Switzerland},
  doi={10.1007/978-3-032-05325-1_14}
}

@article{du2024dper,
  title={{DPER}: Diffusion Prior Driven Neural Representation for Limited Angle and Sparse View CT Reconstruction},
  author={Du, Chenhe and Lin, Xiyue and Wu, Qing and Tian, Xuanyu and Su, Ying and Luo, Zhe and Zheng, Rui and Chen, Yang and Wei, Hongjiang and Zhou, S. Kevin and Yu, Jingyi and Zhang, Yuyao},
  journal={arXiv preprint arXiv:2404.17890},
  year={2024},
  doi={10.48550/arXiv.2404.17890}
}

@article{nikolakakis2024gaspct,
  title={{GaSpCT}: Gaussian Splatting for Novel CT Projection View Synthesis},
  author={Nikolakakis, Emmanouil and Gupta, Utkarsh and Vengosh, Jonathan and Bui, Justin and Marinescu, Razvan},
  journal={arXiv preprint arXiv:2404.03126},
  year={2024},
  doi={10.48550/arXiv.2404.03126}
}

@inproceedings{wu2025dgr,
  title={Discretized Gaussian Representation for Tomographic Reconstruction},
  author={Wu, Shaokai and Lu, Yuxiang and Guo, Yapan and Ji, Wei and Huang, Suizhi and Yang, Fengyu and Sirejiding, Shalayiding and He, Qichen and Tong, Jing and Ji, Yanbiao and Ding, Yue and Lu, Hongtao},
  booktitle={Proceedings of the IEEE/CVF International Conference on Computer Vision},
  pages={25073--25082},
  year={2025}
}

@inproceedings{yang2026ultrags,
  title={{UltraGS}: Real-Time Physically-Decoupled Gaussian Splatting for Ultrasound Novel View Synthesis},
  author={Yang, Yuezhe and Ruan, Qingqing and Cai, Wenjie and Dong, Yudang and Yang, Dexin and Dong, Xingbo and Jin, Zhe and Dai, Yong},
  booktitle={IEEE International Conference on Multimedia and Expo},
  year={2026},
  note={Accepted; arXiv:2511.07743},
  doi={10.48550/arXiv.2511.07743}
}

@inproceedings{song2021ddim,
  title={Denoising Diffusion Implicit Models},
  author={Song, Jiaming and Meng, Chenlin and Ermon, Stefano},
  booktitle={International Conference on Learning Representations},
  year={2021},
  eprint={2010.02502},
  archivePrefix={arXiv}
}

@inproceedings{loshchilov2019decoupled,
  title={Decoupled Weight Decay Regularization},
  author={Loshchilov, Ilya and Hutter, Frank},
  booktitle={International Conference on Learning Representations},
  year={2019},
  eprint={1711.05101},
  archivePrefix={arXiv}
}

@article{wang2004image,
  title={Image quality assessment: From error visibility to structural similarity},
  author={Wang, Zhou and Bovik, Alan C. and Sheikh, Hamid R. and Simoncelli, Eero P.},
  journal={IEEE Transactions on Image Processing},
  volume={13},
  number={4},
  pages={600--612},
  year={2004},
  doi={10.1109/TIP.2003.819861}
}

@inproceedings{ho2020denoising,
  title={Denoising Diffusion Probabilistic Models},
  author={Ho, Jonathan and Jain, Ajay and Abbeel, Pieter},
  booktitle={Advances in Neural Information Processing Systems},
  volume={33},
  pages={6840--6851},
  year={2020}
}

@inproceedings{zhou2019continuity,
  title={On the Continuity of Rotation Representations in Neural Networks},
  author={Zhou, Yi and Barnes, Connelly and Lu, Jingwan and Yang, Jimei and Li, Hao},
  booktitle={Proceedings of the IEEE/CVF Conference on Computer Vision and Pattern Recognition},
  pages={5745--5753},
  year={2019},
  doi={10.1109/CVPR.2019.00589}
}

@inproceedings{vaswani2017attention,
  title={Attention Is All You Need},
  author={Vaswani, Ashish and Shazeer, Noam and Parmar, Niki and Uszkoreit, Jakob and Jones, Llion and Gomez, Aidan N. and Kaiser, Lukasz and Polosukhin, Illia},
  booktitle={Advances in Neural Information Processing Systems},
  volume={30},
  year={2017}
}

\end{document}